\documentclass{article}
\usepackage[preprint]{neurips_2025}
\usepackage{url} 
\usepackage{xcolor} 
\usepackage{amssymb}
\usepackage{amsmath}  
\usepackage{multirow}
\usepackage{hyperref}      
\usepackage{booktabs}      
\usepackage{amsfonts}       
\usepackage{graphicx}
\usepackage{enumitem}
\usepackage{microtype} 
\usepackage{algorithm}
\usepackage{algpseudocode}
\usepackage{hyperref}
\usepackage{tocloft}
\usepackage{minitoc}
\usepackage{footnote}
\usepackage{caption}
\usepackage{float}
\usepackage{bm}
\usepackage{xcolor}

\title{TextOmics-Guided Diffusion for Hit-like Molecular Generation}

\author{
  Hang Yuan$^{1}$ \quad
  Chen Li$^{2}$\thanks{Corresponding author.} \quad
  Wenjun Ma$^{3,4}$\footnotemark[1] \quad
  Yuncheng Jiang$^{1,3}$ \\
  $^1$School of Artificial Intelligence, South China Normal University, China \\
  $^2$D3 Center, The University of Osaka, Japan \\
  $^3$School of Computer Science, South China Normal University, China \\
  $^4$Aberdeen Institute of Data Science and Artificial Intelligence, South China Normal University, China \\
}

\begin{document}
\maketitle
\begin{abstract}
Hit-like molecular generation with therapeutic potential is essential for target-specific drug discovery. However, the field lacks heterogeneous data and unified frameworks for integrating diverse molecular representations. To bridge this gap, we introduce \texttt{TextOmics}, a pioneering benchmark that establishes one-to-one correspondences between omics expressions and molecular textual descriptions. \texttt{TextOmics} provides a heterogeneous dataset that facilitates molecular generation through representations alignment. Built upon this foundation, we propose \texttt{ToDi}, a generative framework that jointly conditions on omics expressions and molecular textual descriptions to produce biologically relevant, chemically valid, hit-like molecules. \texttt{ToDi} leverages two encoders (\texttt{OmicsEn} and \texttt{TextEn}) to capture multi-level biological and semantic associations, and develops conditional diffusion (\texttt{DiffGen}) for controllable generation. Extensive experiments confirm the effectiveness of \texttt{TextOmics} and demonstrate \texttt{ToDi} outperforms existing state-of-the-art approaches, while also showcasing remarkable potential in zero-shot therapeutic molecular generation. Sources are available at: https://github.com/hala-ToDi.
\end{abstract}

%=============================================================
\section{Introduction}
\label{sec:intro}
Targeted molecular therapy \cite{min2022molecular} stands as a fundamental pillar of precision medicine, predicated on a refined molecular-level comprehension of disease pathophysiology \cite{zafar2024revolutionizing}. Compared to non-targeted interventions such as chemotherapy \cite{nygren2001cancer}, which may damage healthy tissues, targeted therapies selectively engage pathological mechanisms, effectively reducing side effects \cite{mukherjee2025data}. Such a shift toward a molecule-directed therapeutic paradigm has fundamentally transformed modern drug development and personalized treatment strategies \cite{adhikary2025updated}. Despite its success, targeted molecular therapy still faces critical barriers to broader implementation. Notably, the rational design of hit-like molecules \cite{ma2024rational, ashraf2024hit}, a crucial step in early-stage target-drug discovery, remains challenging due to prolonged development timelines, high costs, and inadequate integration of biological and contextual information \cite{zeng2022deep}.

% Motivations
Recent advances in artificial intelligence (AI) have enabled the incorporation of biologically contextualized information, such as gene expression profiles into molecular generation, accelerating the discovery of hit-like molecules \cite{cheng2024gexmolgen,li2024gxvaes,matsukiyo2024transcriptionally}. Such approaches seek to uncover the complex relationships between cellular phenotypes and chemical structures, driving the generation of hit-like molecules with desirable pharmacological properties \cite{du2024machine}. However, the biological system is inherently specific, and the chemical space of drug-like molecules is vast and diverse. Consequently, generating molecules that are precisely aligned with specific cellular contexts and underlying biological mechanisms poses a significant challenge, limiting the full therapeutic potential of molecular generation. For example, relying on a single data type, either biological context (e.g., gene expression profiles \cite{li2024gx2mol}) or structural information (e.g., molecular textual descriptions \cite{gong2024text}), fails to capture the complexity of human pathophysiology \cite{zeng2022deep}. These strategies often overlook the contextual intricacies of biological systems, thereby hindering the further advancement of molecular generation. Moreover, despite the abundant knowledge of bioactive molecules, publicly available datasets rarely support the effective integration of heterogeneous representations \cite{mukherjee2025data} or robust frameworks for guided molecular generation \cite{jiang2025network}. Datasets that link omics data and textual descriptions to induce molecular generation remain scarce.

To bridge these gaps, we introduce \texttt{TextOmics}, a self-developed dataset for generating hit-like molecules using heterogeneous data. \texttt{TextOmics} links inducing molecules to biological states through omics data and molecular textual descriptions. Building upon this, we propose \texttt{ToDi} ( \texttt{\underline{T}ext\underline{O}mics}-guided \underline{Di}ffusion), a generative framework that integrates biological data to guide the generation of hit-like molecules. \texttt{ToDi} consists of three core modules: \romannumeral 1) an omics encoder (\texttt{OmicsEn}) for capturing bio-molecular interactions from omics data; \romannumeral 2) a text encoder (\texttt{\texttt{TextEn}}) for extracting semantic embeddings from molecular textual descriptions;  and \romannumeral 3) a diffusion-based generator (\texttt{\texttt{DiffGen}}) that refines molecular structures by conditioning the generation process on the integrated outputs of \texttt{OmicsEn} and \texttt{\texttt{TextEn}}. By combining \texttt{TextOmics} and \texttt{ToDi}, our framework generates biologically informed, context-aware hit-like molecules. The main contributions are as follows:
\begin{itemize}[left=0pt]
\item \textbf{First-of-its-Kind Dataset:} We introduce \texttt{TextOmics}, an open-source dataset that integrates omics data, molecular textual descriptions, and corresponding induced molecular representations, setting a new benchmark for biologically informed hit-like molecular generation.

\item \textbf{Heterogeneity-Aware Framework:} We propose \texttt{ToDi}, an innovative generative framework that integrates heterogeneous data such as omics expressions and molecular textual descriptions via induced molecules, bridging the gap toward hit-like molecular generation.

\item \textbf{State-of-the-Art Performance:} \texttt{ToDi} outperforms existing state-of-the-art (SOTA) approaches targeting molecular generation, demonstrating superior performance across various experiments and showcasing remarkable potential in zero-shot generation of therapeutic candidates.
\end{itemize}

%=============================================================
\section{Related Work}
\label{sec:related}
\textbf{Molecular Generation under Cellular Contexts.} Omics data, such as gene expression profiles, offer insights into gene expression dynamics within human cells, reflecting a wide range of physiological and pathological states \cite{du2024advances,yang2025spatial}. ExpressionGAN \cite{mendez2020novo} pioneered the use of generative adversarial networks \cite{goodfellow2014generative} for generating hit-like molecules conditioned on gene expression signatures by aligning simplified molecular input line entry system (SMILES)-based chemical representations \cite{weininger1988smiles}. Similarly, TRIOMPHE \cite{kaitoh2021triomphe} identifies source molecules whose induced transcriptional profiles most correlate with a target signature, and generates candidates via latent space sampling from a variational autoencoder (VAE) \cite{kingma2014auto}. However, both rely on indirect associations due to the absence of explicit one-to-one correspondences between expression signatures and inducing molecules, limiting biological fidelity and generative robustness.

To address such limitations, GxVAEs \cite{li2024gxvaes}, the current SOTA model, establishes direct associations between expression signatures and corresponding inducing molecules by jointly learning latent representations through dual VAEs.  Extensions such as GxRNN \cite{matsukiyo2024transcriptionally} and HNN2Mol \cite{li2024gx2mol} further incorporate SMILES-based representations using recurrent neural networks (RNNs) \cite{zaremba2014recurrent} and hybrid VAE–RNN architectures. However, the strict syntax and limited expressiveness of SMILES strings constrain chemical validity. More critically, the exclusive emphasis on cellular context neglects the broader biological and semantic heterogeneity inherent in real-world therapeutic scenarios, underscoring the need for integrative, multi-dimensional generative frameworks.

\textbf{Molecular Generation from Textual Descriptions.} Molecular textual descriptions (e.g., functional group annotations and mechanisms of action) provide rich semantic cues for specifying molecular attributes \cite{van2023functional,jung2025interpretable}. Text2Mol \cite{edwards2021text2mol}, the first model to align textual descriptions with molecular structures, introduced a Transformer architecture integrated with graph convolutional embeddings, offering a systematic framework for text-conditioned molecular generation. More recently, large language model-based approaches such as T5 \cite{10.5555/3455716.3455856} and MolT5 \cite{edwards-etal-2022-translation} have emerged, leveraging semantic knowledge from large-scale biomedical corpora to generate molecules with desired properties. However, their autoregressive nature imposes sequential constraints that hider generative flexibility, particularly when exploring the vast chemical space.

The advent of diffusion-based controllable text generation model \cite{li2022diffusion} presents a promising alternative. TGM \cite{gong2024text}, a SMILES-based approach, adopts a two-phase diffusion process to separately perform semantic conditioning and syntax correction, ultimately generating molecular structures. However, due to the syntactic fragility and non-uniqueness of SMILES, the current SOTA TGM requires additional corrective mechanisms to ensure chemical validity. Recently, self-referencing embedded strings (SELFIES) \cite{krenn2020self} has addressed the inherent syntactic invalidity of SMILES, especially in representing complex molecular structures. Despite this advantage, SELFIES often yield more verbose sequences, potentially increasing the learning complexity during generation \cite{krenn2022selfies}.

\textbf{Molecular Generation with Heterogeneous Data.} 
Recent advancements in molecular generation emphasize the integration of heterogeneous data sources, such as omics data, chemical structures, and biomedical annotations, to better capture the intricate nature of biological systems \cite{jiang2025network,zhang2025integration}. Unlike methods that focus solely on molecular representations, models like SiamFlow \cite{tan2023target} and GexMolGen \cite{cheng2024gexmolgen} have incorporated auxiliary graph structures or biological information, though their generative processes remain largely chemistry-oriented. More recent developments, MolGene-E \cite{ohlan2025molgene} enhances this paradigm by aligning omics-derived features with molecular semantics through contrastive learning \cite{hjelm2018learning}, making it possible to generate molecules informed by cellular contexts. However, these frameworks encounter challenges such as coarse-grained representation integration and a lack of explicit mechanisms for semantic control. 

In this study, we introduce \texttt{TextOmics}, a groundbreaking benchmark heterogeneous dataset that establishes a one-to-one linkage between omics data, molecular textual descriptions, and their corresponding induced molecules. Furthermore, we present \texttt{ToDi}, a unified framework within \texttt{TextOmics} that facilitates the generation of hit-like candidates by integrating heterogeneous data and leveraging vocabulary-aware remapped SELFIES representations to enable robust generation.

%=============================================================

\section{Methodology}
\label{sec:framework}

\begin{figure}[t]
\centering
\includegraphics[width=1.0\linewidth]{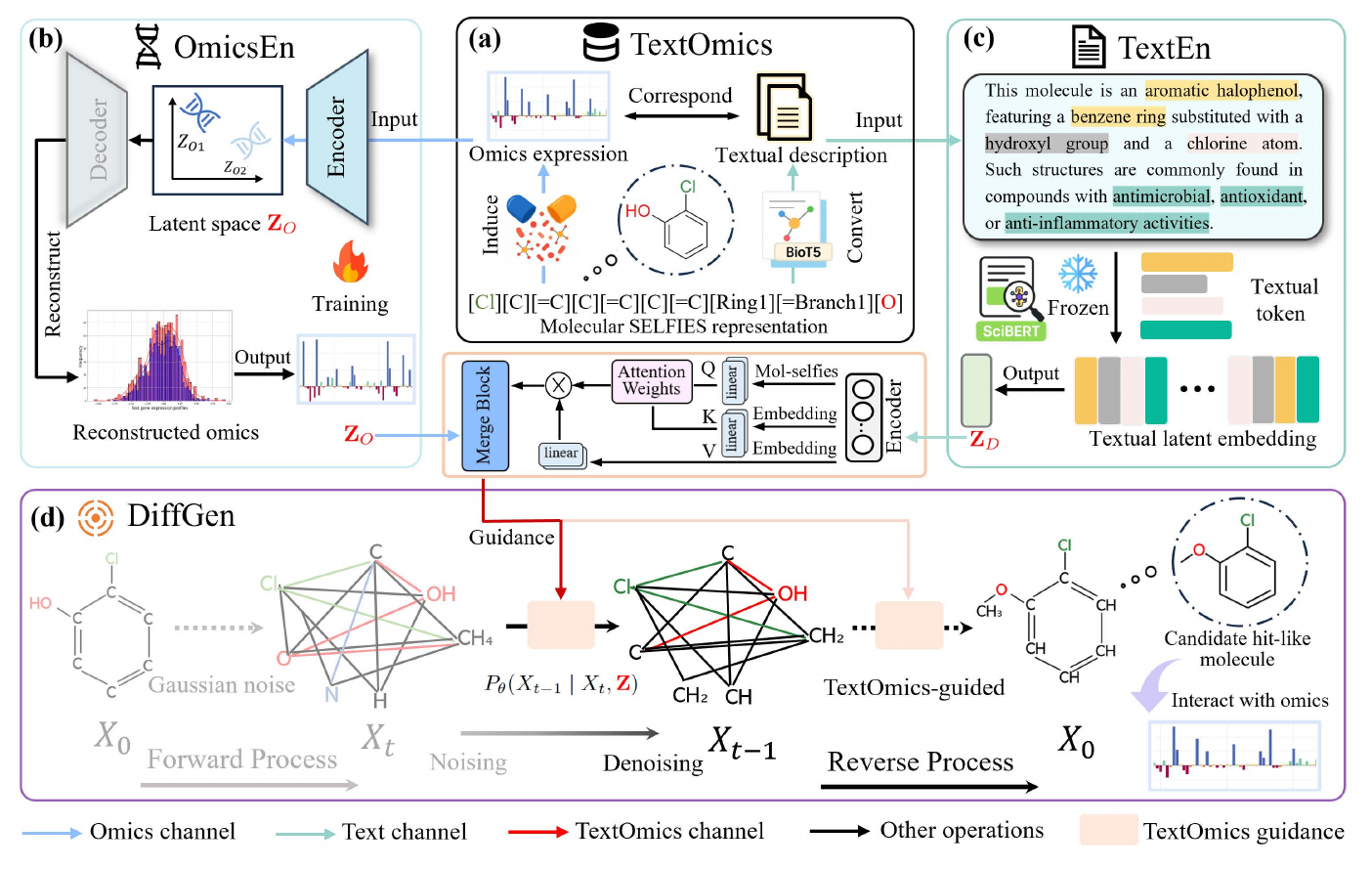}
\caption{Overview of the \texttt{TextOmics} heterogeneous dataset and the \texttt{ToDi} framework for hit-like molecular generation. 
(a) A molecule from \texttt{TextOmics} induces a specific omics expression and is paired with a corresponding molecular textual description, with SELFIES serving as a bridge between these two heterogeneous representations. (b) \texttt{OmicsEn} extracts features from the omics expression and projects them into a latent space to obtain the reconstructed embedding $\mathbf{Z}_O$. Simultaneously, \texttt{TextEn} processes the molecular textual description to acquire semantic features, resulting in the text embedding $\mathbf{Z}_D$. Then, \texttt{ToDi} integrates $\mathbf{Z}_D$ and $\mathbf{Z}_O$ via a cross-attention mechanism to form a joint \texttt{TextOmics}-guided representation $\mathbf{Z}$. (d) Finally, \texttt{DiffGen}, a diffusion-based generator, is developed to produce molecular SELFIES representations conditioned on $\mathbf{Z}$, aiming to generate molecules that exhibit functional interactions with the corresponding induced omics expression.}
\label{fig:overview}
\vspace*{-1\baselineskip} 
\end{figure}

Figure \ref{fig:overview} illustrates our framework: (a) \texttt{TextOmics} consists of three heterogeneous data, including omics expressions reflecting biological context, molecular textual descriptions conveying semantic knowledge, and SELFIES representations encoding chemical structures. SELFIES serves as a bridging representation that enables one-to-one alignment between omics expression and textual descriptions. (b) \texttt{OmicsEn} encodes biological signals from omics expression via a VAE encoder, yielding latent embeddings $\mathbf{Z}_O$ that capture phenotype-level drug response features. (c) \texttt{TextEn} extracts semantic embeddings $\mathbf{Z}_D$ from molecular descriptions using a frozen SciBERT \cite{beltagy-etal-2019-scibert}, providing chemically grounded guidance. (d) \texttt{DiffGen} performs conditional molecular generation via a diffusion process jointly guided by $\mathbf{Z}_O$ and $\mathbf{Z}_D$, integrating biological and semantic conditions to generate syntactically valid, biologically contextualized, hit-like molecules.
%=======================================
%======================================
\subsection{TextOmics: Heterogeneous Data Benchmark}
\label{subsec:TextOmics}
We generate omics-level molecular responses by exposing molecules to specific cellular environments, thereby obtaining corresponding omics expressions. BioT5 \cite{pei-etal-2023-biot5} is then employed to convert SELFIES representations into corresponding textual descriptions. To ensure the accuracy and reliability of such descriptions, all outputs are manually verified by domain experts in chemistry. Once both heterogeneous data are constructed, SELFIES serves as induced molecular representations that effectively link omics expressions and textual descriptions. SELFIES is a robust molecular string representation that encodes chemical structures using a rule-based syntax, guaranteeing syntactic and chemical validity. Compared to SMILES, SELFIES offers enhanced grammatical robustness, making it particularly suitable for diffusion-based generation. Details on SELFIES are in Appendix~\ref{appendix_A.1}.

%=======================
\subsection{ToDi: TextOmics-Guided Generative Framework}
\label{subsec:todi}

\textbf{Omics Expression Encoding.} 
Omics expression reflects the specific internal environment of biological cells. Given an omics expression $\mathbf{\textit{\textbf{E}}} \in \mathbb{R}^K$ representing $K$ gene values, only a biologically meaningful subset of genes typically exhibits expression responses to a given molecule. \texttt{OmicsEn} employs a VAE encoder to capture these biologically meaningful latent representations $\mathbf{Z}_O$, which can be reconstructed back to original omics expression $\mathbf{\textit{\textbf{E}}}$ by the VAE decoder. Specifically, the encoder network approximates the posterior distribution $q_\theta(\mathbf{Z}_O|\mathbf{\textit{\textbf{E}}})$ over the latent variable $\mathbf{Z}_O$, while the decoder reconstructs the input profile $\mathbf{\textit{\textbf{E}}}$ through 
likelihood function $p_\phi(\mathbf{\textit{\textbf{E}}}|\mathbf{Z}_O)$. Subsequently, the model is trained by maximizing the evidence lower bound of the marginal likelihood.
\begin{equation}
\mathcal{L}_{\text{Omics}}(\bm \theta, \bm\phi, \mathbf{\textit{\textbf{E}}}, \mathbf{Z}_O,  \beta) = \mathbb{E}_{q_\theta(\mathbf{Z}_O|\mathbf{\textit{\textbf{E}}})}[\log p_\phi(\mathbf{\textit{\textbf{E}}}|\mathbf{Z}_O)] - \beta \cdot \mathrm{KL}(q_\theta(\mathbf{Z}_O|\mathbf{\textit{\textbf{E}}}) \| p_\phi(\mathbf{Z}_O)),
\end{equation}
where ${\bm\theta}$ and ${\bm\phi}$ denote the parameters of the encoder and decoder, respectively, $\beta$ is a scaling factor that controls the strength of the KL divergence \cite{joyce2011kullback} term, and $\mathbb{E [\ \cdot\ ]}$ is an expectation operation. During inference, $\mathbf{Z}_O$ is sampled via the reparameterization trick $\mathbf{Z}_O = \mu(\mathbf{\textit{\textbf{E}}}) + \sigma(\mathbf{\textit{\textbf{E}}}) \cdot \epsilon$, where $\mu$ and $\sigma$ represent the mean value and standard deviation in the Gaussian distribution, respectively, and $\epsilon \sim \mathcal{N}(0, \mathbf{I})$.
These latent vectors $\mathbf{Z}_O$ encode biologically relevant features of the omics expression $\mathbf{\textit{\textbf{E}}}$ and serve as conditional omics guidance for molecular generation in subsequent modules.

\textbf{Textual Description Encoding.} Textual description encapsulates the chemical structural information of drug molecules. Given a molecular textual description $\mathbf{C}$,  it is tokenized and padded/truncated to a fixed maximum length $L_d$. Only specific descriptive terms, such as `ether', `acid', `ester',  which are a subset of the textual description $\mathbf{C}$, reflect the chemical and structural characteristics of the molecules. To capture the contextual semantics of these terms, donoted as $\mathbf{w}\subset\mathbf{C}$, the textual sample $\mathbf{C}$ is converted into an ID token matrix by a corresponding attention mask belongs to $ \{0,1\}^{L_d}$. Then for any $\mathbf{I}_i$ belongs to the ID token matrix,  where $i\in \mathcal{M}$ represents the input description with the $i$-th token replaced by a [MASK] token and $\mathcal{M}$ denotes the set of masked token positions. Finally, it is fed into the SciBERT operating in evaluation mode and the loss function is defined as 
\begin{equation}
\mathcal{L}_{\text{Text}} = - \sum_{i \in \mathcal{M}} \log P(\mathbf{w} \mid \mathbf{I}_i).
\end{equation}
The encoder’s final output is denoted as $\mathbf{Z}_D$, which corresponding to the full feature of contextualized hidden states from the molecular textual description $\mathbf{C}$, serving as conditional textual guidance for the generation of hit-like molecules.

\textbf{TextOmics-Guided Diffusion Generation.} Integrating heterogeneous data provides abundant and complementary information that enhances molecular generation. \texttt{DiffGen} serves as the core generative module in \texttt{ToDi}, responsible for generating molecular stings under the joint guidance of omics and textual conditions through a conditional diffusion process. To address the overlong sequence issue commonly associated with SELFIES, we introduce a vocabulary-aware sequence remapping strategy that compresses redundant SELFIES tokens while preserving molecular semantics. 
Formally, each SELFIES element $S_i \in \mathcal{V}$ is mapped to a unique index $v_i \in \{0, 1, \dots, |\mathcal{V}| - 1\}$ via a predefined vocabulary mapping function $\phi: \mathcal{V} \rightarrow \mathbb{N}$, where $\mathcal{V}$ is the complete SELFIES token vocabulary. The resulting SELFIES token index $R \subset \mathcal{V}$, derived from the canonical chemical structure order in SELFIES, serving as the input to \texttt{DiffGen}.

At each denoising timestep $t$, \texttt{DiffGen} receives the token indices $R$ and embeds them into representations $ \mathbf{X}_t$. To incorporate SELFIES embedding $\mathbf{X}_t$ and timestep $t$, we further add positional encodings $\mathrm{PE}$ and a timestep embedding $\mathrm{TE}_t$ corresponding to the current diffusion step $t$. These components are summed and passed through a Transformer encoder, where the molecular embedding $\mathbf{X}_t$ serves as the query and the textual context $\mathbf{Z}_D$, extracted from \texttt{TextEn}, acts as the cross-attention condition. The omics embedding $\mathbf{Z}_O$ from \texttt{OmicsEn} is then broadcast across the sequence dimension $L$ and is fused with the $\mathbf{Z}_D$ through token-level concatenation.
\begin{equation}
\mathbf{Z} = \mathcal{T}\left(\mathbf{X}_t + \mathrm{PE} + \mathrm{TE}_t,\; \mathbf{Z}_D \right) \oplus \mathbf{Z}_O,
\end{equation}
where $\mathcal{T}(\cdot)$ denotes the Transformer encoder with cross-attention over $\mathbf{Z}_D$, producing semantically enriched hidden states $\hat{\mathbf{Z}}_t$, which encode the intermediate representation of the molecule at step $t$ under textual guidance. $\oplus$ indicates concatenation along the hidden dimension $H$. The fused representation \( \mathbf{Z} \) serves as the conditioning input to \texttt{DiffGen}, which estimates the conditional distribution \( P_\theta(X_{t-1} \mid X_t, \mathbf{Z}) \) in the denoising process. The final output $\hat{\mathbf{X}}_0$, obtained at the end of the reverse diffusion process, constitutes the predicted molecular embedding used in the training objective. The total training objective combines structural and token-level reconstruction with biological alignment to jointly guide molecular generation.
\begin{equation}
\mathcal{L}_{\text{Total}} = \|\hat{\mathbf{X}}_0 - \mathbf{X}_0\|^2 + \mathcal{L}_{\text{NLL}}(\hat{\mathbf{X}}_0) + \lambda \cdot  \cos(\hat{\mathbf{Z}}_t,\; \mathbf{Z}_O),
\end{equation}

where \(\|\cdot\| \) minimizes the reconstruction error between the predicted $\hat{\mathbf{X}}_0$ and ground-truth molecular embedding $\mathbf{X}_0$, \( \mathcal{L}_{\text{NLL}} \) enforces sequence-level token correctness, and \( \cos(\cdot) \) promotes alignment between the semantic representation $\hat{\mathbf{Z}}_t$ and omics-derived biological context $\mathbf{Z}_O$. The hyperparameter \( \lambda \) controls the strength of omics and textual guidance $\mathbf{Z}$. Details of $\lambda$ selection and its impact on model performance are documented in the Table \ref{table:metrics_l}.

%=============================================================
\section{Experiments}
\label{sec:exp}
\vspace*{-1.5\baselineskip} 
\begin{table}[h]
\setlength{\tabcolsep}{1.6pt}
\caption{Dataset statistics (left) and comparison of data settings with existing SOTA models (right).}
\label{table:statistics}
\begin{minipage}[t]{0.7\textwidth} 
\resizebox{0.98\textwidth}{!}{
\begin{tabular}{l|ccccccccc}\toprule
\texttt{TextOmics} & \#Data & MolLen & OmicsDim & OmicsVar & TextLen & MolWeight & \#Ring & \#Aromatic \\\midrule
ChemInduced & 13755 & 250 & 978 & 0.62 & 62 & 425 & 3.50 & 2.00 \\
TargetPerturb & 31630 & 249 & 978 & 0.91 & 78 & 397 & 3.67 & 2.92 \\
DiseaseSign & 5 & 183 & 884 & 0.01 & 48 & 330 & 2.80 & 1.00 \\\bottomrule
\end{tabular}
}
\end{minipage}%
\begin{minipage}[t]{0.3\textwidth} 
\resizebox{1\textwidth}{!}{
\begin{tabular}{l|ccccc}\toprule
Type & Omics & Text & Pledge & Zero-shot \\\midrule
GxVAEs & \checkmark &  & & \checkmark\\
TGM & & \checkmark  & &\\
\texttt{ToDi} & \checkmark & \checkmark & \checkmark  & \checkmark\\\bottomrule
\end{tabular}
}
\hfill
\end{minipage}
\vspace{-0.2em}
{\scriptsize \S~\#Data: dataset size; MolLen: average SELFIES length; OmicsDim and OmicsVar: dimensionality and mean variance of omics data; TextLen: average length of molecular descriptions; MolWeight, \#Ring, and \#Aromatic: average molecular weight, ring count, and aromatic ring count.}
\vspace*{-1.2\baselineskip} 
\end{table}
%=======================
\subsection{TextOmics Settings}
\texttt{TextOmics} dataset comprises three distinct data types, where SELFIES acts as induced molecular representations that bridge omics expressions and textual descriptions in a one-to-one manner. The left panel of Table \ref{table:statistics} summarizes internal statistics and external comparisons with datasets used in recent SOTA models. The right panel illustrates that ToDi integrates heterogeneous data (omics and text), ensuring the validity of generated molecules and enabling zero-shot capabilities.

\textbf{ChemInduced} was constructed from the LINCS L1000 database \cite{duan2014lincs}, which provides gene expression profiles across 77 human cell lines under chemical perturbations. We specifically used the chemically induced omics expressions of the MCF7 breast cancer cell line in response to 13,755 small molecules. Each molecule is paired with its corresponding molecular textual description.

\textbf{TargetPerturb} was assembled from the LINCS database, comprising gene expression profiles of 77 human cell lines subjected to genetic perturbations, including gene overexpression and knockdown of target proteins. To enrich the biological context, molecular textual descriptions of compounds targeting each protein were also induced. The dataset covers ten cancer-related therapeutic targets: eight with overexpression profiles (AKT1, AKT2, AURKB, CTSK, EGFR, HDAC1, MTOR and PIK3CA) and two with knockdown profiles (SMAD3 and TP53). Details are in Appendix~\ref{appendix_A.2}. 

\textbf{DiseaseSign} was sourced from the CREEDS database \cite{wang2016extraction}, which compiles transcriptomic data from disease samples. Due to the limited availability of approved drugs for Alzheimer's disease and the challenge of generating hit-like molecules, we included Alzheimer's as a case study in \texttt{ToDi}. Omics expressions from multiple patients were averaged to obtain a representative disease-specific signal.

%=======================
\subsection{Evaluation Measures}
%==================
\textbf{Statistical Indicators}. \textit{Validity} measures the proportion of chemically valid molecules, suggesting the legality of the generated candidates. \textit{Uniqueness} is defined as the proportion of non-duplicate molecules among the valid ones, quantifying the diversity of the generation results. \textit{Novelty} reflects the percentage of generated molecules that are absent from the training set, serving as an indicator of the model's tendency to overfit. \textit{Levenshtein} calculates the minimum edit distance between molecular strings and normalizes it to assess their similarity. The calculation details are in Appendix \ref{appendix_E.1}.

\textbf{Molecular Structure.} \textit{Fr\'echet ChemNet Distance} (FCD) quantifies the distributional similarity between generated and reference molecular sets by comparing their latent representations from a pretrained model. \textit{Morgan Tanimoto} captures fine-grained structural similarity by measuring the overlap between ECFP4 fingerprints \cite{rogers2010extended}, which encode atomic-level chemcial environments. \textit{MACCS Tanimoto} assesses structural similarity at a broader level by analyzing predefined substructure-based keys embedded in each molecule \cite{gong2024text}. Details on the calculation are provided in Appendix \ref{appendix_E.2}.

\textbf{Semantic Alignment.} \textit{Hit Ratio} quantifies the alignment between the semantic intent of textual descriptions and the chemical functionality of generated molecules, reflecting the proportion that exhibit the target functional groups implied by the text. This metric is computed by pairing each description with the latent noise of a query molecule and calculating the noise estimation error $|\hat{\epsilon}_\theta - \epsilon|_2^2$ for each pair. Additional details are provided in Algorithm~\ref{alg:hit-ratio}. Note that all metrics are normalized to a range of 0 to 1 for ease of comparison.

%==============================================
\subsection{Omics-Guided Molecular Generation: Can ToDi Surpass Current SOTA?}
\label{sec:RQ1}

\begin{table}[t]
\setlength{\tabcolsep}{5pt}
\centering
\caption{Benchmark results for omics-guided molecular generation on the \texttt{TextOmics} dataset. \texttt{ToDi}\textsubscript{\scriptsize w/o T} refers to generation conditioned solely on omics data. The best and second-best scores for each metric are shown in \textbf{bold} and \underline{underline}, respectively.}
\resizebox{0.9\textwidth}{!}{
\begin{tabular}{l|ccccccc}\toprule
Model & Validity (\%)$\uparrow$ & Uniqueness (\%)$\uparrow$ & Novelty (\%)$\uparrow$ & Levenshtein $\downarrow$ & FCD $\downarrow$ & {Morgan $\uparrow$} & {MACCS $\uparrow$}\\\midrule
TRIOMPHE \cite{kaitoh2021triomphe} & 48.73 & 78.68 & 87.57 & 0.68 & 0.55 & 0.17 & 0.39\\
GxRNN \cite{matsukiyo2024transcriptionally} & 79.23 & 79.35 & 78.00 & 0.67 & 0.52 & 0.20 & 0.45\\
HNN2Mol \cite{li2024gx2mol} & 87.80 & 81.42 & 76.31 & 0.61 & 0.52 & 0.19 & 0.47\\
GxVAEs \cite{li2024gxvaes} & 86.59 & 87.57 & 89.25 & 0.61 & 0.51 & 0.22 & 0.48\\\midrule
\texttt{ToDi}\textsubscript{\scriptsize w/o T} & \textbf{100.0} & \underline{96.73} & \underline{92.23} & \underline{0.59} & \underline{0.49} & \underline{0.28} & \underline{0.51}\\
\texttt{ToDi} & \textbf{100.0} & \textbf{98.45} & \textbf{97.30} & \textbf{0.41} & \textbf{0.33} & \textbf{0.29} & \textbf{0.56}\\\bottomrule
\end{tabular}
}
\label{tab:exp2_res}
\vspace*{-1\baselineskip} 
\end{table} 

Table~\ref{tab:exp2_res} presents the evaluation results of \texttt{ToDi}\textsubscript{\scriptsize w/o T}, the full \texttt{ToDi} framework, and SOTA models on the ChemInduced dataset. \texttt{ToDi}\textsubscript{\scriptsize w/o T} exhibits strong performance, attributable to the syntactic robustness of SELFIES representations combined with the proposed vocabulary-aware sequence remapping strategy. Specifically, it achieves perfect validity (100\%) and improves novelty by 3.0\% relative to the SOTA GxVAEs baseline. These results suggest that the generated hit-like molecules not only maintain chemical validity but also demonstrate enhanced structural diversity. Further improvements are observed upon incorporating molecular textual descriptions, showing the complementary role of semantic guidance. Although \texttt{ToDi}\textsubscript{\scriptsize w/o T} already achieves robust results under omics-only supervision, the full \texttt{ToDi} framework consistently delivers superior performance by effectively integrating both biological and semantic information, thereby enabling more diverse and precise molecular generation. Notably, \texttt{ToDi} achieves an additional 5.1\% increase in novelty over \texttt{ToDi}\textsubscript{\scriptsize w/o T}, surpassing the improvement margin between \texttt{ToDi}\textsubscript{\scriptsize w/o T} and the prior SOTA. This margin illustrates the contribution of textual guidance in generating novel and diverse molecules.

From the structural evaluation perspective, \texttt{ToDi}\textsubscript{\scriptsize w/o T} already outperforms existing baselines across all structural metrics, demonstrating the effectiveness of our framework in capturing biologically meaningful representations solely from omics expressions. Upon incorporating molecular textual descriptions, the full \texttt{ToDi} framework generates hit-like molecules exhibiting enhanced structural similarity to the known molecules. For example, the FCD score is further reduced by 0.16 compared to \texttt{ToDi}\textsubscript{\scriptsize w/o T}, indicating the semantic guidance facilitates the generation of molecules more faithfully aligned with the underlying chemical structure of reference molecules.

Furthermore, Figure~\ref{fig:vae-pca} provides supporting analysis confirming that \texttt{OmicsEn} effectively captures biologically relevant features from omics expressions. Additional evaluation metrics, including \textit{RDK} Tanimoto similarity and \textit{QED} score, along with their detailed results, are presented in Table~\ref{tab:F_6}.

%==============================================
\subsection{Text-Guided Molecular Generation: What Is the Potential of ToDi?}
\label{sec:RQ2}
\definecolor{brightmaroon}{rgb}{0.76, 0.13, 0.28}   % 暗红色（酒红）
\definecolor{bleudefrance}{rgb}{0.19, 0.55, 0.91}  % 深蓝色（藏青）

\begin{figure}[t]
\centering
\includegraphics[width=\linewidth]{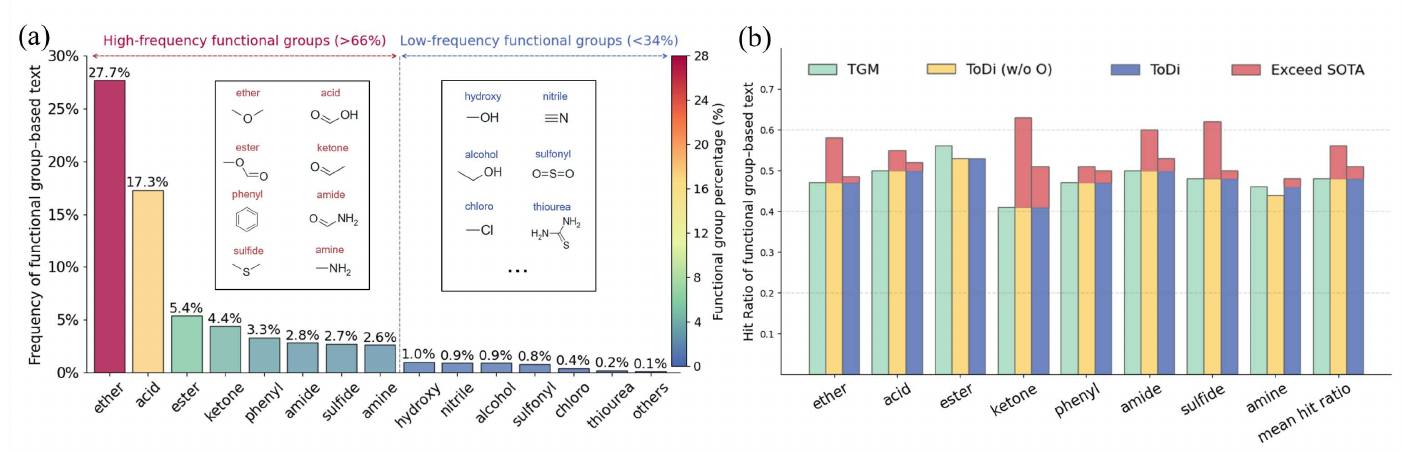}
\caption{Functional group distribution in molecular textual descriptions and corresponding hit ratio performance. (a) Frequency distribution of functional group–based textual descriptions in \texttt{TextOmics}, with the top eight designated as high-frequency (\textcolor{brightmaroon}{red}) and the remainder as low-frequency (\textcolor{bleudefrance}{blue}). (b) Hit ratios of molecules generated under various textual conditions by the SOTA baseline TGM, \texttt{ToDi}\textsubscript{\scriptsize w/o O} (text-only guided generation), and the full \texttt{ToDi} framework.}
\label{fig:exp2_res}
\vspace*{-1\baselineskip} 
\end{figure}

Figure~\ref{fig:exp2_res} (a) provides a statistical overview of molecular textual descriptions in the \texttt{TextOmics} dataset, categorized by functional groups. The top eight functional groups, including ether and acid, collectively account for over 66\% of all occurrences, whereas less frequent groups like hydroxy are comparatively underrepresented. This distribution reflects the inherent semantic composition of biomedical molecular descriptions and provides foundation for functional group-based evaluations. Additional distributional details are shown in Figure~\ref{fig:word}.

Figure~\ref{fig:exp2_res} (b) compares hit ratios across different settings. Under the text-only condition, \texttt{ToDi}\textsubscript{\scriptsize w/o O} achieves a higher average hit ratio than the SOTA baseline TGM, outperforming it in six of the top eight most frequent functional group categories. This indicates the strong capability of \texttt{ToDi}\textsubscript{\scriptsize w/o O} in generating hit-like molecules guided solely by textual descriptions. Upon incorporating omics data, the overall hit ratio of the full \texttt{ToDi} slightly decreases, as expected, due to the additional biological constraints introduced. Notably, \texttt{ToDi} outperforms the SOTA TGM in seven of the top eight high-frequency scenarios and even surpasses \texttt{ToDi}\textsubscript{\scriptsize w/o O}, demonstrating its enhanced capacity to generate biologically coherent molecules under joint \texttt{TextOmics} guidance. Hit ratio results under randomly sampled textual descriptions are provided in Figure~\ref{fig:hit-ratio-desc}.

Moreover, we conducted additional experiments on the ChEBI-20 \cite{edwards2021text2mol} benchmark to ensure a fair comparison. Similar to the notable gains in statistical metrics achieved by \texttt{ToDi}\textsubscript{\scriptsize w/o T}, \texttt{ToDi}\textsubscript{\scriptsize w/o O} also demonstrates consistently superior results across relevant statistical indicators under textual-only guidance, indicating superior semantic fidelity in the generated hit-like molecules. Structural evaluation confirms enhanced alignment with reference compounds, showing the crucial role of description-guided generation in preserving chemical semantics and structural plausibility. Comprehensive evaluation metrics and corresponding quantitative results are summarized in Table~\ref{table:text}.

%=========================================
\subsection{TextOmics-Guided Molecular Generation: How Well Does ToDi Generalize?}
\label{sec:RQ3}

\texttt{ToDi} exhibits a distinctive capability to integrate heterogeneous data for hit-like molecular generation. To assess its generalizability, we transferred \texttt{ToDi} from ChemInduced to TargetPerturb dataset. In contrast, SOTA baselines, limited to single-source inputs, were evaluated under the same setting. This setup illustrates the cross-task transferability of \texttt{ToDi} and its robustness under multi-source scenarios.

Table~\ref{tab:exp3_res} shows that \texttt{ToDi} consistently surpasses all baselines across ten target-specific ligand generation tasks. Notably, it improves uniqueness by 10.3\% and reduces the Levenshtein distance by 0.13, compared to the strongest competitor. These results indicate that \texttt{ToDi} generates syntactically valid, diverse, and novel molecules, effectively tailored to specific therapeutic targets. In terms of structural evaluation, \texttt{ToDi} consistently delivers strong performance across all ten therapeutic targets. For instance, on AURKB target protein, it reduces the FCD to 0.44, outperforming the closest baseline by 0.32. \texttt{ToDi} also achieves the highest Tanimoto similarity under both Morgan and MACCS fingerprints on most targets, including perfect MACCS scores (1.00) for six out of ten proteins. 
On the finest-grained Morgan Tanimoto similarity evaluation, \texttt{ToDi} surpasses the strongest baseline by as much as 0.43, representing a relative improvement of 91\%. These results underscore \texttt{ToDi}’s ability to generate structurally coherent molecules aligned with known ligands across diverse biological contexts. This performance stems from its effective integration of omics-derived biological features and textual semantics, which jointly guide the generation process toward structurally and biologically plausible candidates. Additional metrics reported in Table~\ref{tab:extra-metrics} further validate \texttt{ToDi}’s capacity to produce physiochemically aligned molecules. 
%============================
\begin{table}[t]
\centering
\setlength{\tabcolsep}{3pt}
\caption{Comparison of \texttt{ToDi} with current SOTA approaches across ten target proteins.}
\resizebox{1.0\textwidth}{!}{
\begin{tabular}{l|ccc|ccc|ccc|ccc|ccc|ccc|ccc}\toprule
\multirow{2}{*}{Target} & \multicolumn{3}{c|}{Validity (\%)$\uparrow$} & \multicolumn{3}{c|}{Uniqueness (\%)$\uparrow$} & \multicolumn{3}{c|}{Novelty (\%)$\uparrow$} & \multicolumn{3}{c|}{Levenshtein $\downarrow$} & \multicolumn{3}{c|}{FCD $\downarrow$} & \multicolumn{3}{c|}{Morgan $\uparrow$} & \multicolumn{3}{c}{MACCS $\uparrow$}\\
 & GxVAEs & TGM & \texttt{ToDi} & GxVAEs & TGM & \texttt{ToDi} & GxVAEs & TGM & \texttt{ToDi} & GxVAEs & TGM & \texttt{ToDi} & GxVAEs & TGM & \texttt{ToDi}  & GxVAEs & TGM & \texttt{ToDi} & GxVAEs & TGM & \texttt{ToDi}\\
\midrule
AKT1   & 88.00 & 72.00 & \textbf{100.0} & 88.60 & 75.29 & \textbf{98.91} & 100.0 & 95.00 & \textbf{100.0} & 0.89 & 0.90 & \textbf{0.76} & 0.75 & 0.77 & \textbf{0.45} & \textbf{0.85} & 0.47 & \underline{0.70} & 1.00 & 0.81 & \textbf{1.00}\\
AKT2   & 89.00 & 71.47 & \textbf{100.0} & 91.00 & 77.84 & \textbf{98.76} & 98.80 & 96.12 & \textbf{100.0} & 0.89 & 0.83 & \textbf{0.72} & 0.78 & 0.79 & \textbf{0.46} & 0.43 & 0.39 & \textbf{0.76} & 0.78 & 0.75 & \textbf{1.00}\\
AURKB  & 89.00 & 76.23 & \textbf{100.0} & 93.30 & 78.43 & \textbf{99.12} & 98.80 & 100.0 & \textbf{100.0} & 0.83 & 0.85 & \textbf{0.73} & 0.80 & 0.76 & \textbf{0.44} & 0.47 & 0.46 & \textbf{0.90} & 0.78 & 0.79 & \textbf{1.00}\\
CTSK   & 91.00 & 82.94 & \textbf{100.0} & 94.50 & 70.01 & \textbf{98.25} & 98.80 & 99.50 & \textbf{100.0} & 0.89 & 0.97 & \textbf{0.85} & 0.73 & 0.70 & \textbf{0.42} & 0.38 & 0.39 & \textbf{0.64} & 0.75 & 0.72 & \textbf{0.96}\\
EGFR   & 89.00 & 75.70 & \textbf{100.0} & 93.30 & 80.02 & \textbf{98.40} & 100.0 & 99.00 & \textbf{100.0} & 0.73 & 0.83 & \textbf{0.72} & 0.85 & 0.80 & \textbf{0.47} & 0.74 & 0.50 & \textbf{0.77} & 0.87 & 0.80 & \textbf{1.00}\\
HDAC1  & 78.00 & 70.61 & \textbf{100.0} & 96.20 & 78.11 & \textbf{99.48} & 97.30 & 98.20 & \textbf{100.0} & 0.90 & 0.84 & \textbf{0.74} & 0.74 & 0.75 & \textbf{0.47} & 0.55 & 0.46 & \textbf{0.65} & 0.80 & 0.74 & \textbf{0.97}\\
MTOR   & 91.00 & 79.62 & \textbf{100.0} & 93.40 & 83.33 & \textbf{99.61} & 98.80 & 99.90 & \textbf{100.0} & 0.78 & 0.87 & \textbf{0.77} & 0.73 & 0.83 & \textbf{0.50} & 0.52 & 0.42 & \textbf{0.70} & 0.83 & 0.82 & \textbf{0.94}\\
PIK3CA & 92.00 & 80.37 & \textbf{100.0} & 93.50 & 81.17 & \textbf{99.14} & 97.70 & 94.52 & \textbf{100.0} & 0.82 & 0.88 & \textbf{0.77} & 0.92 & 0.89 & \textbf{0.61} & 0.35 & 0.42 & \textbf{0.47} & 0.78 & 0.79 & \textbf{0.92}\\
SMAD3  & 86.00 & 80.89 & \textbf{100.0} & 91.90 & 66.00 & \textbf{96.86} & 98.70 & 99.80 & \textbf{99.98} & 0.73 & 0.71 & \textbf{0.66} & 0.67 & 0.69 & \textbf{0.37} & \textbf{0.98} & 0.83 & \underline{0.88} & 0.94 & 0.96 & \textbf{1.00}\\
TP53   & 85.00 & 81.00 & \textbf{100.0} & 96.50 & 70.81 & \textbf{97.75} & 98.80 & 70.66 & \textbf{99.95} & 0.77 & 0.72 & \textbf{0.63} & 0.72 & 0.75 & \textbf{0.43} & 0.76 & 0.73 & \textbf{0.88} & 0.94 & 0.86 & \textbf{1.00}\\\bottomrule
\end{tabular}
}
\label{tab:exp3_res}
\end{table}
%=================================
\begin{figure}[t]
\centering
\includegraphics[width=1.0\textwidth]{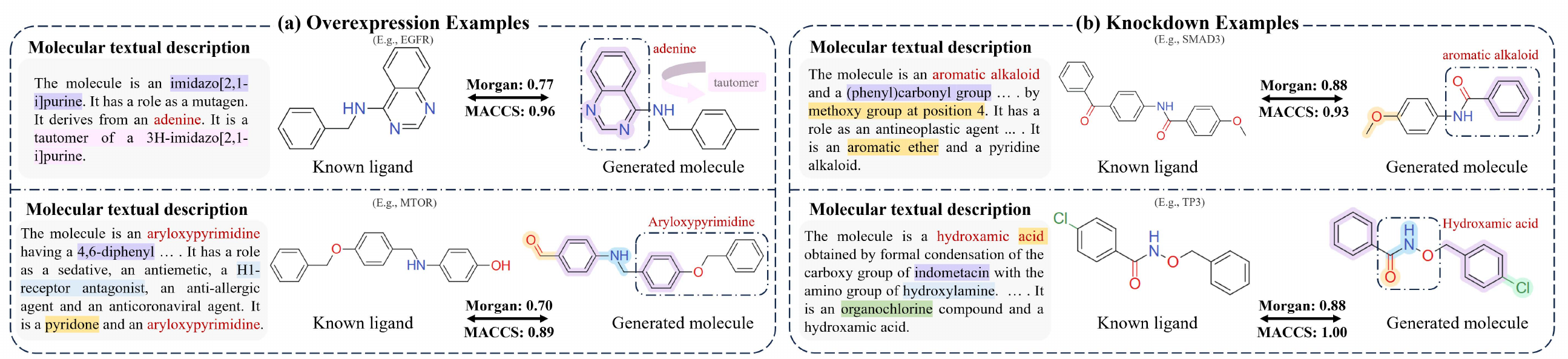}
\caption{Detailed illustration of \texttt{ToDi}'s performance in terms of biological similarity and textual hit under both overexpression and knockdown conditions.}
\label{fig:exp3_fig}
\vspace*{-1\baselineskip} 
\end{figure}
%=============================

Figure~\ref{fig:exp3_fig} demonstrates that \texttt{ToDi} accurately captures functional groups specified in textual prompts and generates hit-like molecules exhibiting high Tanimoto similarity to known ligands. Beyond broad category matching, \texttt{ToDi} effectively encodes fine-grained chemical semantics, exemplified by its correct interpretation and generation of tautomers explicitly indicated in the molecular text description. Importantly, consistent performance is observed under both overexpression (e.g., MTOR, EGFR) and knockdown (e.g., TP53, SMAD3) target conditions, supporting the robust generalizability of our framework. Additional results for molecules exhibiting high structural similarity to the ligands are presented in Figure~\ref{fig:ligands}.

Furthermore, we conducted comprehensive ablation studies on \texttt{ToDi} to assess the individual contributions of its key components. Specifically, we examined two ablated variants: \texttt{ToDi}\textsubscript{\scriptsize w/o T}, which excludes molecular textual descriptions, and \texttt{ToDi}\textsubscript{\scriptsize w/o O}, which excludes omics expressions. These two variants were evaluated alongside the full \texttt{ToDi} framework that integrates heterogeneous data. As shown in Table~\ref{ablation_study}, both ablated variants still outperform existing SOTA models under single-source settings, demonstrating the effectiveness of each data modality. Notably, the full \texttt{ToDi} framework consistently achieves the best overall performance, highlighting the synergistic advantage of integrating heterogeneous sources for hit-like molecular generation.

%=======================
\subsection{Therapeutic Molecular Generation: Can ToDi Perform Zero-shot for Specific Diseases?}

In real-world clinical scenarios, especially for diseases with a limited repertoire of approved therapeutics such as Alzheimer's disease, diagnosis and treatment decisions often rely heavily on patient-reported symptoms and the clinical expertise of physicians \cite{adhikary2025updated}. While this approach remains feasible, current methodologies lack the capacity to jointly exploit both textual symptom descriptions and rich patient-specific omics profiles for direct personalized therapeutic design. To address this gap, our goal is to leverage \texttt{ToDi} to generate biologically informed and structurally relevant drug-like candidate molecules, thereby enabling the tailored design of therapeutics for specific diseases.

We apply the pretrained \texttt{ToDi} framework to the DiseaseSign dataset to generate biologically grounded hit-like molecules. To more closely approximate real-world clinical scenarios, we replace structured molecular descriptions with diagnostic narratives derived from Alzheimer's disease patients, enabling the evaluation of \texttt{ToDi}'s zero-shot generation capability. The generated candidates are benchmarked against those produced by GxVAEs. To assess structural relevance to existing treatments, we compute the Morgan and MACCS Tanimoto similarity scores between the generated molecules and currently approved drugs for Alzheimer's disease.

\begin{figure}[t]
\centering
\includegraphics[width=1.0\linewidth]{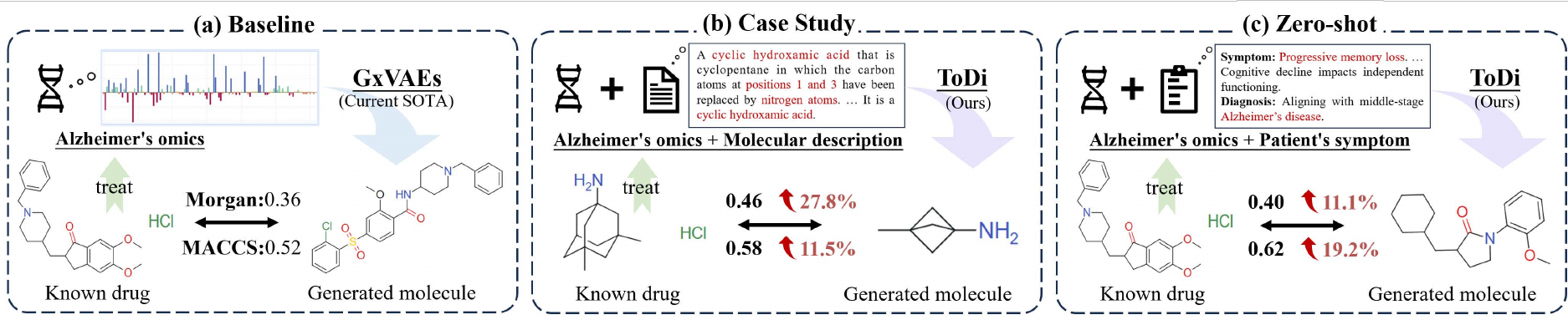}
\caption{Evaluation of \texttt{ToDi}'s performance in zero-shot generation of biologically meaningful hit-like molecules, guided by molecular textual descriptions or patient symptom narratives.}
\label{fig:sample}
\vspace*{-0.5\baselineskip} 
\end{figure}

Figure \ref{fig:sample} presents the representative examples of hit-like molecules generated by \texttt{ToDi} under two distinct guidance settings: molecular textual description and patient symptom narratives. Under molecular textual description guidance, \texttt{ToDi} successfully produced molecular structures closely resembling the approved drug Memantine hydrochloride. The generated candidates achieved Morgan and MACCS Tanimoto similarity scores of 0.46 and 0.58, corresponding to relative improvements of 27.8\% and 11.5\% compared to the SOTA GxVAEs baseline, respectively. These results demonstrate \texttt{ToDi}’s robust capacity to capture both chemical semantic information from molecular textual descriptions and disease-specific biological signals relevant to Alzheimer's pathology.

In the zero-shot setting, where molecular textual descriptions were replaced by patient symptom narratives, \texttt{ToDi} generated candidate molecules exhibiting greater similar to the known drug Donepezil hydrochloride, paralleling the results observed with GxVAEs. This result supports our hypothesis that, given symptom narratives of Alzheimer's patients are absent from the \texttt{TextOmics} corpus, \texttt{ToDi} places increased reliance on omics expressions for molecular design, thereby producing molecules with stronger biological relevance. Despite the challenging conditions, \texttt{ToDi} outperformed GxVAEs by 11.1\% and 19.2\% in similarity scores. These results further substantiate \texttt{ToDi}’s ability to integrate heterogeneous data sources and demonstrate its potential to generate disease-tailored therapeutic hit-like molecules, demonstrating remarkable adaptability in zero-shot contexts.

%=========================================
\section{Conclusion}
We introduce \texttt{TextOmics}, the first benchmark that bridges omics expressions, molecular text descriptions, and induced molecular SELFIES representations for hit-like molecular generation. Additionally, we present \texttt{ToDi}, a unified generative framework that integrates biological and semantic contexts from \texttt{TextOmics} to guide conditional generation of molecules via diffusion. Comprehensive experiments validate the efficacy of \texttt{TextOmics} and reveal that \texttt{ToDi} surpasses current SOTA approaches, exhibiting exceptional potential for zero-shot therapeutic molecular generation.

\texttt{ToDi} has two primary limitations: i) the generated molecules have not undergone experimental validation, and ii) the current evaluation is confined to case studies on Alzheimer's disease, lacking comprehensive investigation of rare diseases such as Gaucher disease. Future work will explore clinical relevance, including in vitro screening and experimental validation of generated candidates to advance practical drug discovery applications. Additionally, we plan to establish dedicated rare disease datasets and conduct validations to generate therapeutic candidates for rare diseases. 

%\paragraph{Broader impacts.}
%\texttt{TextOmics} introduces a pioneering benchmark that integrates omics expression data with molecular textual descriptions, providing a comprehensive framework for biologically grounded of drug discovery. Leveraging this foundation, the proposed \texttt{ToDi} framework enables biologically and semantically guided hit-like molecular generation, which offers a novel paradigm for designing therapeutics that are chemically relevant and biologically contextualized. This heterogeneous integration has the potential to accelerate early-stage drug discovery, especially in scenarios with limited labeled data or unexplored targets, such as rare diseases or emerging pathogens.

%Despite its potential, the deployment of generative methods in drug design comes with inherent challenges and responsibilities. The generation of novel compounds must be carefully validated for safety and efficacy before clinical application. Additionally, access to patient-derived omics data must be handled with rigorous attention to privacy and ethical standards.  By openly releasing the \texttt{TextOmics} benchmark and the \texttt{ToDi} framework, we hope to foster transparent and reproducible research and to catalyze innovation in data-integrated, precision therapeutics.

%=============================================================
\bibliographystyle{unsrt}
\bibliography{refs}

%=============================================================
%=============================================================
%=============================================================
\clearpage
\appendix
\setcounter{secnumdepth}{4}
\numberwithin{figure}{section}
\numberwithin{table}{section}
\numberwithin{algorithm}{section}
\onecolumn
\begin{center}
\LARGE \textbf{Appendix}
\end{center}

\vspace{1.2em}
\noindent
\vspace{1.0em}
\textbf{A \quad Dataset Details} \dotfill~\pageref{appendix_A} \\
\vspace{1.0em}
\hspace{1.5em} A.1 \quad SELFIES \dotfill~\pageref{appendix_A.1} \\
\vspace{1.0em}
\hspace{1.5em} A.2 \quad Omics Data \dotfill~\pageref{appendix_A.2} \\
\vspace{1.0em}
\hspace{1.5em} A.3 \quad Molecular Textual Description \dotfill~\pageref{appendix_A.3} \\
\vspace{1.0em}

\noindent
\vspace{1.0em}
\textbf{B \quad Baseline} \dotfill~\pageref{appendix_B} \\
\vspace{1.0em}
\hspace{1.5em} B.1 \quad The Principles for Selecting Baseline Methods \dotfill~\pageref{appendix_B.1} \\
\vspace{1.0em}
\hspace{1.5em} B.2 \quad Baseline Methods \dotfill~\pageref{appendix_B.2} \\
\vspace{1.0em}

\noindent
\vspace{1.0em}
\textbf{C \quad Experiment Setting} \dotfill~\pageref{appendix_C} \\
\vspace{1.0em}
\hspace{1.5em} C.1 \quad Data Partitioning \dotfill~\pageref{appendix_C.1} \\
\vspace{1.0em}
\hspace{1.5em} C.2 \quad Hyperparameter Setting \dotfill~\pageref{appendix_C.2} \\
\vspace{1.0em}

\noindent
\vspace{1.0em}
\textbf{D \quad Ablation Study} \dotfill~\pageref{appendix_D} \\
\vspace{1.0em}

\noindent
\vspace{1.0em}
\textbf{E \quad Measures Details} \dotfill~\pageref{appendix_E} \\
\vspace{1.0em}
\hspace{1.5em} E.1 \quad Statistical Indicators \dotfill~\pageref{appendix_E.1} \\
\vspace{1.0em}
\hspace{1.5em} E.2 \quad Molecular Structure \dotfill~\pageref{appendix_E.2} \\
\vspace{1.0em}
\hspace{1.5em} E.3 \quad Semantic Alignment \dotfill~\pageref{appendix_E.3} \\
\vspace{1.0em}

\noindent
\vspace{1.0em}
\textbf{F \quad Evaluation Details} \dotfill~\pageref{appendix_F} \\
\vspace{1.0em}
\hspace{1.5em} F.1 \quad Omics-Guided Molecular Generation \dotfill~\pageref{appendix_F.1} \\
\vspace{1.0em}
\hspace{1.5em} F.2 \quad Text-Guided Molecular Generation \dotfill~\pageref{appendix_F.2} \\
\vspace{1.0em}
\hspace{1.5em} F.3 \quad TextOmics-Guided Molecular Generation \dotfill~\pageref{appendix_F.3}

%=============================================================
% \section{Appendix}
% \addcontentsline{toc}{section}{Appendix Contents}
% \tableofcontents

\renewcommand{\thesubsection}{\Alph{subsection}}
% \renewcommand{\thesubsubsection}{\thesubsection.\arabic{subsubsection}}
 % 附录表格编号为 A.1, A.2...
% \setcounter{table}{0}
%===============================================
\newpage
\subsection{Dataset Details}
\label{appendix_A}
To complement the main text, we provide additional details regarding the construction, characteristics, and composition of the heterogeneous data used in our proposed framework. These include the molecular SELFIES, the biological omics datasets, and the textual descriptions aligned to molecular semantics. Such elaboration facilitates a deeper understanding of the \texttt{TextOmics} benchmark and its foundational role in guiding semantically and biologically grounded molecular generation.
%===============================================
\renewcommand{\thesubsubsection}{A.\arabic{subsubsection}}
\subsubsection{SELFIES}

\renewcommand{\thetable}{A.\arabic{table}} 
\renewcommand{\thefigure}{A.\arabic{figure}}
\label{appendix_A.1}
SELFIES \cite{krenn2020self} representation is a robust, semantically constrained molecular string format designed to overcome the syntactic fragility of traditional representations such as SMILES. Introduced by Krenn et al. in 2020, SELFIES ensures that every possible string corresponds to a valid molecular graph. This is achieved by enforcing a formal grammar that embeds chemical valency rules directly into the string-generation process, thereby eliminating invalid molecules at the syntactic level. This property makes SELFIES particularly suitable for generative models, where arbitrary string outputs must remain chemically meaningful. In addition, the token-level discreteness of SELFIES facilitates efficient integration with sequence-based architectures and discrete latent variable models.
%===============================================

\begin{figure}[h]
\centering
\includegraphics[width=1.0\linewidth]{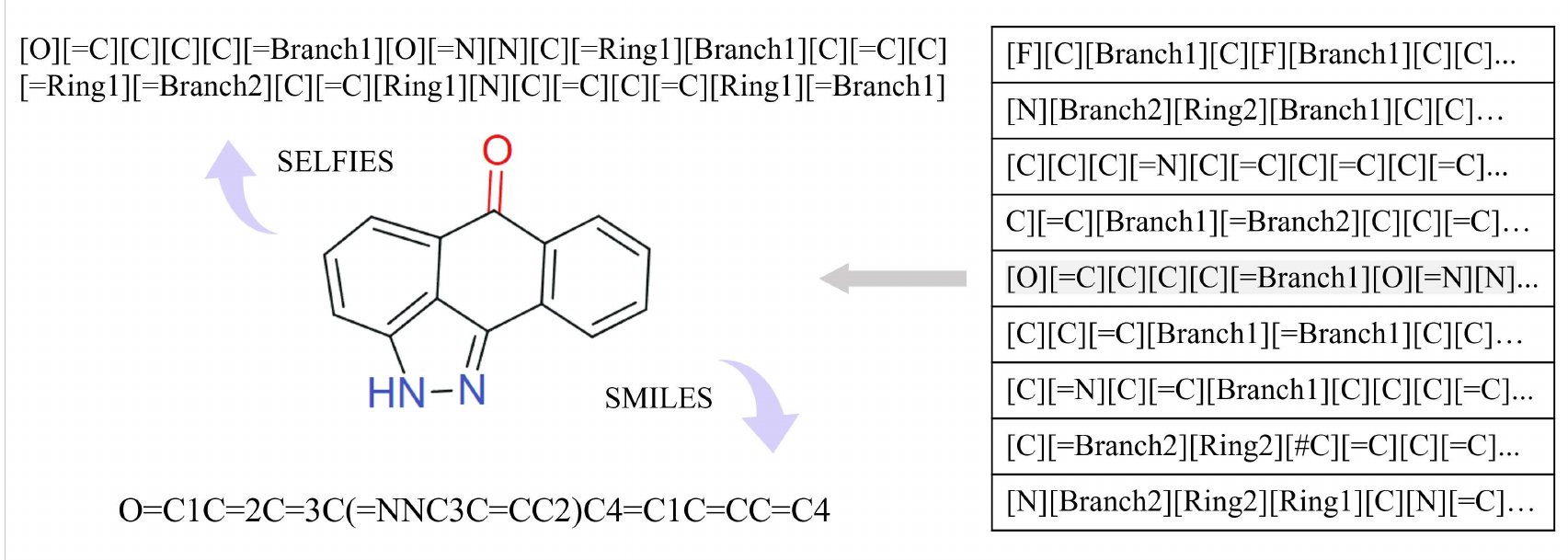}
\caption{The specific representation styles of SEFLIES and their SMILES.}
\label{fig:selfies}
\vspace*{-0.5\baselineskip} 
\end{figure}
%======================
\paragraph*{Advantages of SELFIES.} 
\textbf{Robustness:} The most critical advantage of SELFIES is that it guarantees chemical validity — every SELFIES string maps to a valid molecular graph. \textbf{Compatibility:} SELFIES can be easily converted to and from SMILES using existing toolkits, facilitating interoperability. \textbf{Token-level Semantics:} Tokens carry chemically meaningful information, which helps in learning context-aware molecular structures. \textbf{Generate efficiently:} SELFIES reduces the rate of invalid outputs in models such as VAEs, GANs, and diffusion-based molecular generators.

%===============================================
\paragraph*{Limitations of SELFIES.} 
\textbf{Verbosity:} SELFIES representations are often significantly longer than their SMILES counterparts due to the explicit encoding of structural constraints, which can increase model learning complexity and memory usage. \textbf{Reduced Uniqueness:} Because SELFIES uses a grammar-based expansion, different SELFIES representations may correspond to the same molecule, potentially increasing redundancy. \textbf{Less Human-Readable:} Compared to SMILES, SELFIES representations are more difficult to interpret manually, which can be a limitation in interpretability-focused applications.

%===============================================
\paragraph*{Comparison with SMILES.} 
In contrast to SMILES, which encodes molecules using ASCII characters in a linear notation based on a depth-first traversal of the molecular graph, SELFIES represents molecules as a sequence of discrete tokens drawn from a finite vocabulary (e.g., [C], [=O], [Branch1], etc.). Details are illustrated in Figure \ref{fig:selfies}. These tokens are interpreted using a context-free decoder that incrementally builds a molecular graph while checking valency constraints at each step. This design guarantees that any string formed by valid SELFIES tokens yields a chemically valid molecule, regardless of the string's composition or length. Such robustness makes SELFIES particularly advantageous for generative models, where output stability and chemical feasibility are critical. Moreover, its symbolic token structure aligns well with sequence-based architectures such as Transformers, enabling more effective learning and manipulation of molecular syntax. Table \ref{table:selfies} presents a comparison with SMILES across five key dimensions.

\begin{table}[t]
\caption{Comparative analysis between SMILES and SELFIES}
\label{table:selfies}
\centering
\renewcommand{\arraystretch}{1.2}
\begin{tabular}{l|cc}\toprule
Property & SMILES & SELFIES \\\midrule
String Validity   & Not guaranteed & 100\% guaranteed \\
String Length   & Compact and concise & Generally longer and more verbose \\
Interpretability  & Highly human-readable & Machine-focused; less human-readable \\
Model Training   & Requires constraints/post-filters & No need for filtering or masking \\
Generate Results   & May produce invalid strings & Always generates valid molecules \\\bottomrule
\end{tabular}
\end{table}
%===============================================
\subsubsection{Omics Data}
\label{appendix_A.2}
We employ three categories of gene expression datasets to support different molecular generation scenarios, each reflecting a distinct biological context.  All datasets are preprocessed to retain the top 978 landmark genes from the LINCS L1000 platform, and further standardized before experiment. All publicly available datasets (e.g., LINCS, CREEDS) used in this study are released under open-access or research use–friendly licenses. License information is provided at the original data source links.

\paragraph*{ChemInduced.}
This dataset is derived from the LINCS L1000 project \cite{duan2014lincs} and contains transcriptomic profiles of human cell lines treated with various small molecules. Each sample represents a unique drug–cell line–dose–time combination. In our experiments, we focus on the MCF7 cell line and extract profiles associated with 13,755 compounds. These chemically induced profiles serve as the foundation for learning the transcriptomic consequences of compound perturbation, enabling drug-to-profile alignment and guiding conditional molecular generation.

%===============================================
\paragraph*{TargetPerturb.}
Target perturbation data is also sourced from the LINCS L1000 database, involving gene expression responses following knockdown or overexpression of specific protein targets. Specifically, RAC-$\alpha$ serine/threonine-protein kinase (AKT1), RAC-$\beta$ serine/threonine-protein kinase (AKT2), Aurora B kinase (AURKB), cysteine synthase A (CTSK), epidermal growth factor receptor (EGFR), histone deacetylase 1 (HDAC1), mammalian target of rapamycin (MTOR), phosphatidylinositol 3-kinase catalytic subunit (PIK3CA), decapentaplegic homolog 3 (SMAD3), and tumor protein p53 (TP53). These profiles provide insight into the regulatory impact of molecular targets on the cell’s transcriptomic state, and are used for generating molecules conditioned on desired target-level effects. This setup supports target-driven molecule design.

%===============================================
\paragraph*{DiseaseSign.}
To simulate real-world therapeutic design, we utilize CREEDS (Crowd Extracted Expression of Differential Signatures) \cite{wang2016extraction}, which aggregates curated gene expression signatures for various diseases. Each disease profile is a population-level average representing the transcriptional deviation between disease and normal conditions. We focus on Alzheimer’s disease due to the scarcity and design difficulty of effective targeted drugs, in order to evaluate the \texttt{ToDi}’s ability to generate disease-relevant hit molecules.

%===============================================
\subsubsection{Molecular Textual Description}
\label{appendix_A.3}

\begin{figure}[t]
\centering
\includegraphics[width=0.8\linewidth]{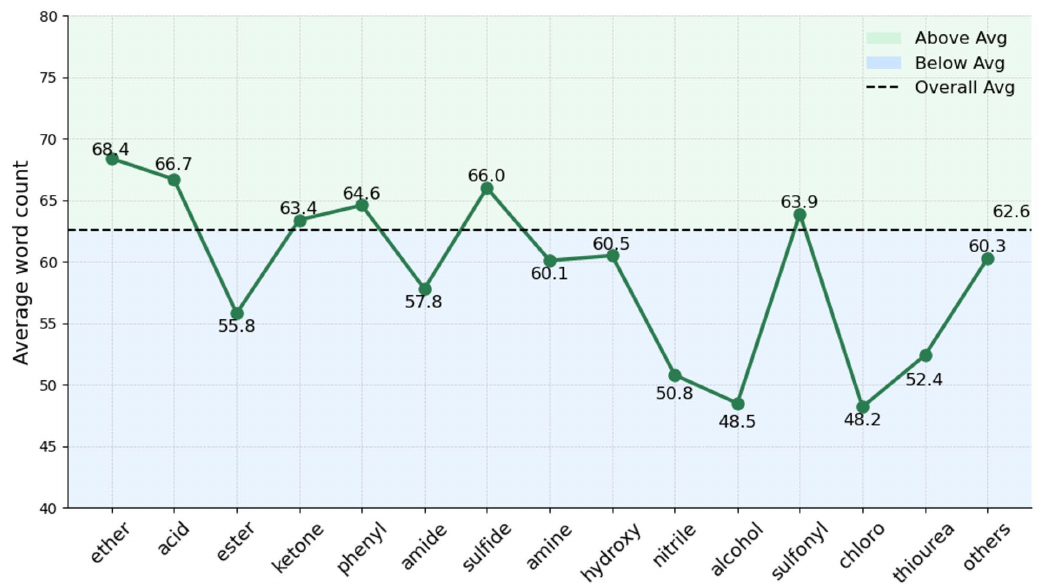}
\caption{The average count of words described in the molecular text represented by the functional groups in the ChemInduced test set.}
\label{fig:word}
\vspace*{-0.5\baselineskip} 
\end{figure}

\paragraph*{Data Sources.}
The molecular textual descriptions in the TextOmics dataset are generated to provide semantic insights into the structural and functional attributes of drug molecules.  Each description is derived from its corresponding SELFIES-based representation using a pretrained BioT5 model fine-tuned on biomedical corpora.  To ensure accuracy and interpretability, all textual outputs are manually verified by domain experts with backgrounds in chemistry and pharmacology.

\paragraph*{Data Description.} These descriptions typically contain information about functional groups, substructures, and mechanisms of action, reflecting both low-level chemical features and high-level bioactivity annotations.  The average sequence length across the dataset is 62–78 tokens, depending on the data subset (e.g., ChemInduced or TargetPerturb). Figure \ref{fig:word} shows the average number of words described by the molecular text represented by the functional group in the ChemInduced test set.  Importantly, each textual description is uniquely aligned with both its corresponding omics expression profile and molecular structure, enabling fine-grained heterogeneous data integration.

%参考文献引入过来
\paragraph*{Introduction Effect.} This textual data serves a dual purpose: it enriches the semantic conditioning space during molecular generation and enables interpretability by aligning chemical language with biological response profiles.  Such alignment empowers the \texttt{ToDi} framework to learn meaningful associations between textual semantics and molecular structures, ultimately guiding the generation of biologically relevant and chemically valid hit-like molecules.

%===============================================
\renewcommand{\thesubsubsection}{B.\arabic{subsubsection}}
\subsection{Baseline}
\label{appendix_B}
\texttt{ToDi} is the first to explore the use of heterogeneous data, like \texttt{TextOmics}, to guide the generation of hit-like molecules. There exists no direct precedent for fully multimodal baselines. Accordingly, we collect baselines from prior methods based on either omics data or textual descriptions.

%===============================================
\subsubsection{Principles for Selecting Baseline Methods}
\label{appendix_B.1}
\paragraph*{Modal Matching.} To disentangle the contributions of textual and omics data, we include variants of \texttt{ToDi} with either source ablated (i.e., \texttt{ToDi}\textsubscript{w/o O} and \texttt{ToDi}\textsubscript{w/o T}), thereby enabling controlled comparisons across different input conditions.

\paragraph*{Task Generalizability.} Models such as TRIOMPHE and GxVAEs , which are representative of gene-expression-based conditional generation methods, are included to reflect the omics-guided generation landscape.

\paragraph{Architecture Alignment.} We adopt Transformer-based models (e.g., MolT5) and diffusion-based model TGM that serve as strong text-conditioned generation baselines, consistent with the backbone design of \texttt{ToDi}.

\paragraph*{Fair Comparison.} All baseline models are retrained or adapted to the same dataset splits and evaluation settings, ensuring comparability in terms of input representations, vocabulary size, and molecule length constraints.

Through this principled selection, we ensure that the comparative results meaningfully reflect the unique advantages of \texttt{ToDi}’s cross-modal design, rather than implementation-specific artifacts.
%===============================================
\subsubsection{Baseline Methods}
\label{appendix_B.2}
\paragraph*{Omics-based.} Methods based on gene expression profiles include TRIOMPHE \cite{kaitoh2021triomphe}, which decodes screened source molecules using a VAE model; GxRNN \cite{matsukiyo2024transcriptionally}, which directly feeds gene expression profiles as conditional inputs to a single RNN to generate molecules with similar expression signatures; GxVAEs \cite{li2024gxvaes}, which maps gene expression and molecular structures into a shared latent space using a dual-VAE architecture to generate hit-like molecules; and HNN2Mol \cite{li2024gx2mol}, which combines a VAE-based encoder and a condition-aware LSTM generator to produce biologically consistent SMILES.
\paragraph*{Textual-based.} Methods based on textual descriptions include Transformer, which employs the standard encoder–decoder architecture and is directly trained on molecular textual description datasets; T5-Base \cite{10.5555/3455716.3455856}, a pretrained sequence-to-sequence model that can be fine-tuned on molecular text to generate molecular sequences; MolT5-Base \cite{edwards-etal-2022-translation}, which is initialized from T5 and further pretrained to capture domain-specific knowledge of molecules and semantic connotations, followed by fine-tuning for molecular generation; and TGM \cite{gong2024text}, which adopts a two-phase diffusion process to generate valid SMILES representations guided by textual descriptions.

%===============================================
\renewcommand{\thesubsubsection}{C.\arabic{subsubsection}}
\renewcommand{\thetable}{C.\arabic{table}} 
\renewcommand{\thefigure}{C.\arabic{figure}}
\subsection{Experiment Setting}
\label{appendix_C}

\subsubsection{Data Partitioning}
\label{appendix_C.1}
We randomly split the ChemInduced dataset into a 9:1 ratio for the training and evaluation of \texttt{ToDi}\textsubscript{\scriptsize w/o T} and \texttt{ToDi}. This experimental setting is consistently adopted throughout subsequent experiments for both training and preliminary evaluation of the models. The TargetPerturb dataset is used as a case study to generate ligand candidates for ten specified molecular targets, aiming to assess the ability of \texttt{ToDi} to produce biologically grounded and semantically meaningful molecules under the guidance of \texttt{TextOmics}.
In contrast, the DiseaseSign dataset is designed to approximate real-world conditions and is used to explore the zero-shot generation capabilities of \texttt{ToDi} in scenarios with no target-specific training molecular textual samples.

\subsubsection{Hyperparameter Setting}
\label{appendix_C.2}
\texttt{ToDi} is implemented with Pytorch framework accelerated by NVlDIA GeForce RTX 3090 GPU, Inteli7-9700K CPU. Python 3.8 is used as the programming language. In \texttt{OmicsEn}, both the encoder and decoder consist of three feedforward layers with dimensions of 512, 256, and 128, respectively. The learning rate and dropout are set to $1\mathrm{e}{-4}$ and 0.2. In \texttt{TextEn}, the maximum sequence length for tokenized SELFIES representations is set to 258, and the trainable token embedding dimension is 32. A frozen encoder with an embedding size of 768 is used to extract features from textual descriptions. In \texttt{DiffGen}, the number of diffusion steps is set to $T = 2000$, with a learning rate of $1\mathrm{e}{-4}$ and a dropout rate of 0.1. To further enhance the sampling efficiency of remapped SELFIES, a uniform skip-sampling strategy is adopted during the reverse diffusion.

To ensure the reliability of our relevance alignment strategy, we conducted a systematic ablation over the weighting coefficient $\lambda$ across a range of cosine similarity thresholds. As shown in Table \ref{table:metrics_l}, $\lambda=0.3$ achieves the best overall balance across multiple evaluation metrics, including BLEU, Levenshtein, FCD, and Tanimoto-based similarities (MACCS, RDK, Morgan), while maintaining perfect validity and high uniqueness. These results empirically support our choice of $\lambda=0.3$ as an optimal setting for guiding molecular generation under \texttt{TextOmics} conditions.

\begin{table}[t]
\caption{\texttt{Textomics}-guided molecular generation results on the ChemInduced test set under different relevance levels.}
\label{table:metrics_l}
\centering
\resizebox{1.0\textwidth}{!}{
\renewcommand{\arraystretch}{1.2}
\begin{tabular}{l|ccccccccc}\toprule
Values & Validity(\%)$\uparrow$ & Uniqueness(\%)$\uparrow$ & Novelty(\%)$\uparrow$ & BLEU$\downarrow$ & FCD$\downarrow$ & Levenshtein$\downarrow$ & MACCS$\uparrow$ & RDK$\uparrow$ & Morgan$\uparrow$ \\\midrule
$\lambda=0.0$ & 100.0 & 95.24 & 96.71 & 0.77 & 0.60 & 0.40 & 0.34 & 0.24 & 0.11 \\
$\lambda=0.1$ & 100.0 & 97.99 & 98.95 & 0.82 & 0.52 & 0.37 & 0.42 & 0.25 & 0.14 \\
$\lambda=0.3$ & 100.0 & 98.45 & 97.30 & \textbf{0.84} & \textbf{0.41} & \textbf{0.33} & \textbf{0.56} & \textbf{0.40} & \textbf{0.33} \\
$\lambda=0.5$ & 100.0 & 98.54 & 98.85 & 0.81 & 0.53 & 0.35 & 0.42 & 0.25 & 0.14 \\
$\lambda=0.7$ & 100.0 & 99.74 & 98.97 & 0.82 & 0.54 & 0.36 & 0.43 & 0.25 & 0.14 \\
$\lambda=0.9$ & 100.0 & \textbf{100.0} & \textbf{100.0} & \textbf{0.84} & 0.52 & 0.59 & 0.42 & 0.25 & 0.15 \\
$\lambda=1.0$ & 100.0 & \textbf{100.0} & \textbf{100.0} & 0.83 & 0.53 & 0.35 & 0.43 & 0.25 & 0.15 \\\bottomrule
\end{tabular}
}
\end{table}

\subsection{Ablation Study}
\label{appendix_D}
\renewcommand{\thetable}{D.\arabic{table}} 

To further validate the contributions of each modality, we conduct an ablation study comparing three configurations of \texttt{ToDi}—with and without omics guidance (\texttt{ToDi}\textsubscript{\scriptsize w/o O}) and textual guidance (\texttt{ToDi}\textsubscript{\scriptsize w/o T})—alongside two state-of-the-art baselines: GxVAEs and TGM. As shown in Table \ref{ablation_study}, each individual component in \texttt{ToDi} substantially contributes to overall performance, and their combination yields the strongest results.

\paragraph*{Effectiveness of the Omics Module.}
Conversely, \texttt{ToDi}\textsubscript{\scriptsize w/o T}, guided solely by omics profiles, shows noticeable improvements over GxVAEs, particularly in novelty (92.23\% vs. 89.25\%) and Levenshtein distance (0.59 vs. 0.61). This highlights that omics-derived embeddings encode biologically relevant information that enhances the generation of novel and syntactically coherent structures.

\paragraph*{Effectiveness of the Text Module.}
\texttt{ToDi}\textsubscript{\scriptsize w/o O}, which relies solely on textual descriptions, significantly outperforms TGM across most metrics. It achieves a validity of 100\%, uniqueness of 97.31\%, and novelty of 88.27\%, showing that textual semantics alone can effectively guide the generation of chemically valid and diverse molecules. Compared to TGM, it also reduces the Levenshtein distance from 0.57 to 0.51, demonstrating better sequence-level similarity with ground-truth structures.

\paragraph*{Benefits of TextOmics Joint Guidance.}
When both modalities are integrated, \texttt{ToDi} achieves the best overall performance across all reported metrics. It attains 100\% validity, 98.45\% uniqueness, and 97.30\% novelty—outperforming all other settings. Furthermore, it lowers the Levenshtein distance to 0.41, indicating that molecules generated under joint guidance are not only valid and diverse, but also structurally closer to true compounds. Notably, \texttt{ToDi} achieves the highest structural similarity under both Morgan and MACCS Tanimoto metrics (0.29 and 0.56, respectively), underscoring the effectiveness of multimodal guidance in aligning chemical features with desired properties.

These results clearly demonstrate the complementary nature of omics and textual signals in the \texttt{TextOmics} framework. Each contributes unique semantic or biological context, and their integration in \texttt{ToDi} leads to synergistic improvements in both chemical plausibility and target specificity.

\begin{table}[t]
\setlength{\tabcolsep}{5pt}
\centering
\caption{Ablation study and comparison with state-of-the-art baselines (GxVAEs and TGM) on the \texttt{TextOmics} benchmark. \texttt{ToDi}\textsubscript{\scriptsize w/o T} denotes generation guided solely by omics data, while \texttt{ToDi}\textsubscript{\scriptsize w/o O} uses only textual descriptions. The best and second-best scores for each metric are shown in \textbf{bold} and \underline{underline}, respectively.}
\resizebox{0.9\textwidth}{!}{
\begin{tabular}{l|ccccccc}\toprule
Model & Validity (\%)$\uparrow$ & Uniqueness (\%)$\uparrow$ & Novelty (\%)$\uparrow$ & Levenshtein $\downarrow$ & FCD $\downarrow$ & {Morgan $\uparrow$} & {MACCS $\uparrow$}\\\midrule
GxVAEs \cite{li2024gxvaes} & 86.59 & 87.57 & 89.25 & 0.61 & 0.51 & 0.22 & 0.48\\
TGM \cite{gong2024text} & 80.83 & 76.71 & 90.25 & 0.57 & 0.57 & 0.19 & 0.42\\\midrule
\texttt{ToDi}\textsubscript{\scriptsize w/o T} & \textbf{100.0} & 96.73 & \underline{92.23} & 0.59 & \underline{0.49} & \underline{0.28} & \underline{0.51}\\
\texttt{ToDi}\textsubscript{\scriptsize w/o O} & \textbf{100.0} & \underline{97.31} & 88.27 & \underline{0.51} & 0.50 & 0.25 & 0.49\\
\texttt{ToDi} & \textbf{100.0} & \textbf{98.45} & \textbf{97.30} & \textbf{0.41} & \textbf{0.33} & \textbf{0.29} & \textbf{0.56}\\\bottomrule
\end{tabular}
}
\label{ablation_study}
\vspace*{-1\baselineskip} 
\end{table}

\renewcommand{\thesubsubsection}{E.\arabic{subsubsection}}
\renewcommand{\thealgorithm}{E.\arabic{algorithm}}
\subsection{Measures Details}

\label{appendix_E}
To provide a comprehensive understanding of \texttt{ToDi} framework, this section details the computation procedures for the three categories of evaluation metrics: Statistical Indicators, Molecular Structure, and Semantic Alignment. These measures collectively assess the quality, diversity, and relevance of the generated molecules from different perspectives, offering a holistic view of model performance.
\subsubsection{Statistical Indicators}
\label{appendix_E.1}

We detail the computation of the four statistical indicators used to evaluate the quality, diversity, and generalization of generated molecules. Let the full set of generated molecules be denoted as $\mathcal{G} = \{g_1, g_2, \dots, g_N\}$.

%=====================================================
\paragraph*{Validity.} This metric assesses the proportion of chemically valid molecules within the generated set. A molecule $g_i \in \mathcal{G}$ is considered valid if it passes chemical sanitization checks (e.g., via RDKit or SELFIES decoding). Let $\mathcal{V} \subseteq \mathcal{G}$ denote the subset of valid molecules. The validity is computed as:
\[
\text{Validity} = \frac{|\mathcal{V}|}{|\mathcal{G}|}
\]

\paragraph*{Uniqueness.} This measures the proportion of unique (non-duplicate) molecules among the valid subset $\mathcal{V}$. Let $\text{unique}(\mathcal{V})$ be the set of distinct molecules in $\mathcal{V}$. The uniqueness is defined as:
\[
\text{Uniqueness} = \frac{|\text{unique}(\mathcal{V})|}{|\mathcal{V}|}
\]

\paragraph*{Novelty.} Novelty evaluates the proportion of generated molecules that are not present in the training set $\mathcal{D}_{\text{train}}$. It reflects the model's ability to generate novel candidates beyond memorized samples. The novelty is calculated as:
\[
\text{Novelty} = \frac{|\{g \in \mathcal{G} \mid g \notin \mathcal{D}_{\text{train}}\}|}{|\mathcal{G}|}
\]

\paragraph*{Levenshtein Similarity.} To assess the string-level similarity between generated molecules and their ground-truth references, we compute the normalized Levenshtein distance between each pair $(g_i, r_i)$, where $r_i$ is the reference molecule paired with $g_i$. Let $d(g_i, r_i)$ be the edit distance between the two strings. The normalized similarity is defined as:
\[
\text{LevSim}(g_i, r_i) = 1 - \frac{d(g_i, r_i)}{\max(|g_i|, |r_i|)}
\]
The overall score is averaged across all generated-reference pairs:
\[
\text{Levenshtein Similarity} = \frac{1}{N} \sum_{i=1}^{N} \text{LevSim}(g_i, r_i)
\]

%====================================

\subsubsection{Molecular Structure}
\label{appendix_E.2}
We describe the structural evaluation metrics used to assess how closely the generated molecules resemble known or reference compounds in terms of chemical structure.

%=================================
\paragraph*{Fr\'echet ChemNet Distance.} FCD quantifies the distributional similarity between generated molecules $\mathcal{G}$ and reference molecules $\mathcal{R}$ in a learned chemical feature space. Specifically, it computes the Fréchet distance between two multivariate Gaussians fitted to the activations of a pretrained ChemNet model over both sets. Let $(\mu_g, \Sigma_g)$ and $(\mu_r, \Sigma_r)$ be the means and covariances of the embeddings of the generated and reference molecules, respectively. Then the FCD is calculated as:
\[
\text{FCD} = \|\mu_g - \mu_r\|^2 + \operatorname{Tr}\left(\Sigma_g + \Sigma_r - 2\left(\Sigma_g \Sigma_r\right)^{1/2}\right)
\]

\paragraph*{Morgan Tanimoto.} This metric computes the average Tanimoto similarity between generated and reference molecules using ECFP4 fingerprints, which encode atom-centered topological environments. Let $\mathbf{f}_g$ and $\mathbf{f}_r$ denote the binary ECFP4 fingerprints of a generated molecule and its matched reference, respectively. The pairwise similarity is given by:
\[
\text{Tanimoto}(\mathbf{f}_g, \mathbf{f}_r) = \frac{|\mathbf{f}_g \cap \mathbf{f}_r|}{|\mathbf{f}_g \cup \mathbf{f}_r|}
\]
The final score is averaged over all molecule pairs.

\paragraph*{MACCS Tanimoto.} This metric operates similarly to the Morgan Tanimoto, but instead uses MACCS keys, a fixed-length binary fingerprint based on predefined chemical substructures. Let $\mathbf{m}_g$ and $\mathbf{m}_r$ be the MACCS fingerprints of the generated and reference molecules. The Tanimoto similarity is computed as:
\[
\text{Tanimoto}(\mathbf{m}_g, \mathbf{m}_r) = \frac{|\mathbf{m}_g \cap \mathbf{m}_r|}{|\mathbf{m}_g \cup \mathbf{m}_r|}
\]
This score reflects structural similarity at a coarser, substructure-based level.

%==================================
\subsubsection{Semantic Alignment}
\label{appendix_E.3}

\paragraph*{Hit Ratio.} This metric quantifies the alignment between molecular textual descriptions and the chemical functionality of the generated molecules. Specifically, it evaluates whether the generated candidate reflects the semantic intent—typically encoded as a target functional group—of the input molecular textual description.

Algorithm~\ref{alg:hit-ratio} outlines the full evaluation process. For each textual description, we sample $N$ molecular candidates using the pretrained model $\mathcal{M}$ under a fixed sampling method $\mathcal{S}$ across $T$ denoising steps. Let $\hat{\mathbf{x}}_i$ denote the $i$-th decoded molecule and $g_i$ its corresponding ground-truth functional group label.

A generated molecule $\hat{\mathbf{x}}_i$ is considered a \emph{hit} if it exhibits the target functional group specified in the input description. In practice, we determine this by comparing either:
- the decoded molecule $\hat{\mathbf{x}}_i$ with the functional group keyword parsed from the input text, or
- the associated noise estimation error $\|\hat{\epsilon}_\theta - \epsilon\|_2^2$ against a predefined threshold $\delta$.

Formally, the Hit Ratio is computed as:
\[
\text{Hit Ratio} = \frac{1}{N} \sum_{i=1}^N \mathbb{I} \left[ \text{Match}(\hat{\mathbf{x}}_i, g_i) \;\text{or}\; \|\hat{\epsilon}_\theta^{(i)} - \epsilon^{(i)}\|_2^2 \leq \delta \right]
\]
where $\mathbb{I}[\cdot]$ is the indicator function, and $\delta$ is the average noise error across all generated samples. This metric reflects the proportion of molecules that semantically satisfy the target condition.
%============================================
\begin{center}
\begin{minipage}{0.85\linewidth}
\begin{algorithm}[H]
\small
\caption{\texttt{ToDi} Hit Ratio Evaluation}
\label{alg:hit-ratio}
\begin{algorithmic}[1]
\State \textbf{Inputs:} Pretrained model $\mathcal{M}$, tokenizer $\mathcal{T}$, dataset $\mathcal{D}$, sample number $N$
\State \textbf{Hyperparameters:} batch size $B$, noise steps $T$, sampling method $\mathcal{S}$
\State Load model weights and set $\mathcal{M}$ to eval mode
\State $(\mathbf{D}_\text{state}, \mathbf{D}_\text{mask}) \gets$ extract $N$ samples from $\mathcal{D}$
\State Initialize buffer $\mathcal{R} \gets \emptyset$, \quad $\mathcal{E} \gets \emptyset$
\For{$i = 0$ \textbf{to} $N$ \textbf{step} $B$}
    \State $(\mathbf{x}_i, \epsilon_i) \gets \mathcal{S}(\mathcal{M}, \mathbf{D}_\text{state}[i{:}i{+}B], \mathbf{D}_\text{mask}[i{:}i{+}B])$
    \State $\mathcal{R} \gets \mathcal{R} \cup \mathbf{x}_i$
    \State $\mathcal{E} \gets \mathcal{E} \cup \epsilon_i$
\EndFor
\State \textbf{Decode}: $\hat{\mathbf{x}} \gets \mathcal{T}^{-1}(\arg\max \mathcal{M}(\mathcal{R}))$
\State \textbf{Hit Ratio}: compare $\hat{\mathbf{x}}$ with functional group targets
\State \textbf{Return:} generated molecules $\hat{\mathbf{x}}$ and corresponding noise errors $\mathcal{E}$
\end{algorithmic}
\end{algorithm}
\end{minipage}
\end{center}

%=======================================
\renewcommand{\thesubsubsection}{F.\arabic{subsubsection}}
\renewcommand{\thetable}{F.\arabic{table}}
\renewcommand{\thefigure}{F.\arabic{figure}}
\subsection{Evaluation Details}
\label{appendix_F}
To comprehensively evaluate the effectiveness of ToDi under the TextOmics framework, we conducted an extensive series of experiments and ablation studies. While the main paper presents key findings, this section provides additional details and supplementary results that could not be included in the main text due to space constraints.

\subsubsection{Omics-Guided Molecular Generation}
\label{appendix_F.1}

\paragraph*{Extended Evaluation Metrics.} 
%======================================
\begin{table}[t]
\setlength{\tabcolsep}{5pt}
\centering
\caption{Extended benchmark results of omics-guided molecular generation on \texttt{TextOmics} test set. The best and second-best scores for each metric are highlighted in \textbf{bold} and \underline{underline}, respectively. \texttt{ToDi}\textsubscript{\scriptsize w/o T} denotes generation results guided solely by omics data.}
\resizebox{0.7\textwidth}{!}{
\begin{tabular}{l|ccc|cc}\toprule
Model & MolWeight & \#Ring & \#Aromatic & {RDK $\uparrow$} & {QED $\uparrow$}\\\midrule
TRIOMPHE \cite{kaitoh2021triomphe} & 421 & 3.10 & 2.23 & 0.32 & 0.54\\
GxRNN \cite{matsukiyo2024transcriptionally} & 437 & 3.45 & 2.10 & 0.34 & 0.57\\
HNN2Mol \cite{li2024gx2mol} & 423 & 3.33 & 1.97 & 0.35 & 0.59\\
GxVAEs \cite{li2024gxvaes} & 436 & 3.63 & 2.07 & 0.36 & 0.59\\\midrule
\texttt{ToDi}\textsubscript{\scriptsize w/o T} & 354 & 2.87 & 1.49 & \underline{0.36} & \underline{0.60} \\
\texttt{ToDi} & 370 & 3.25 & 1.75 & \textbf{0.40} & \textbf{0.64}\\\bottomrule
\end{tabular}
}
\label{tab:F_6}
\vspace*{-1\baselineskip} 
\end{table} 
%======================================
Compared to prior omics-guided generation methods, \texttt{ToDi} demonstrates a balanced capacity to generate structurally reasonable and pharmacologically promising molecules. As shown in Table \ref{tab:F_6}, It achieves the highest QED score (0.64) and the best RDK similarity (0.40) among all baselines, indicating superior drug-likeness and alignment with chemical features typically associated with bioactive compounds.

Although \texttt{ToDi} generates molecules with moderately lower molecular weight and fewer aromatic rings than GxVAEs or GxRNN, it preserves structural complexity with an average ring count of 3.25—comparable to existing models. The improvements over \texttt{ToDi}\textsubscript{w/o T} across all metrics, particularly in QED (0.64 vs. 0.60) and aromaticity (1.75 vs. 1.49), suggest that integrating textual guidance provides complementary information beyond omics signals alone.

These findings highlight \texttt{ToDi}'s effectiveness in generating chemically and pharmacologically meaningful molecules under joint omics-text conditions, outperforming structure-only and omics-only baselines across multiple structure-aware metrics.

\paragraph*{Omics Reconstruction Analysis.} 
\begin{figure}[t]
\centering
\includegraphics[width=0.85\linewidth]{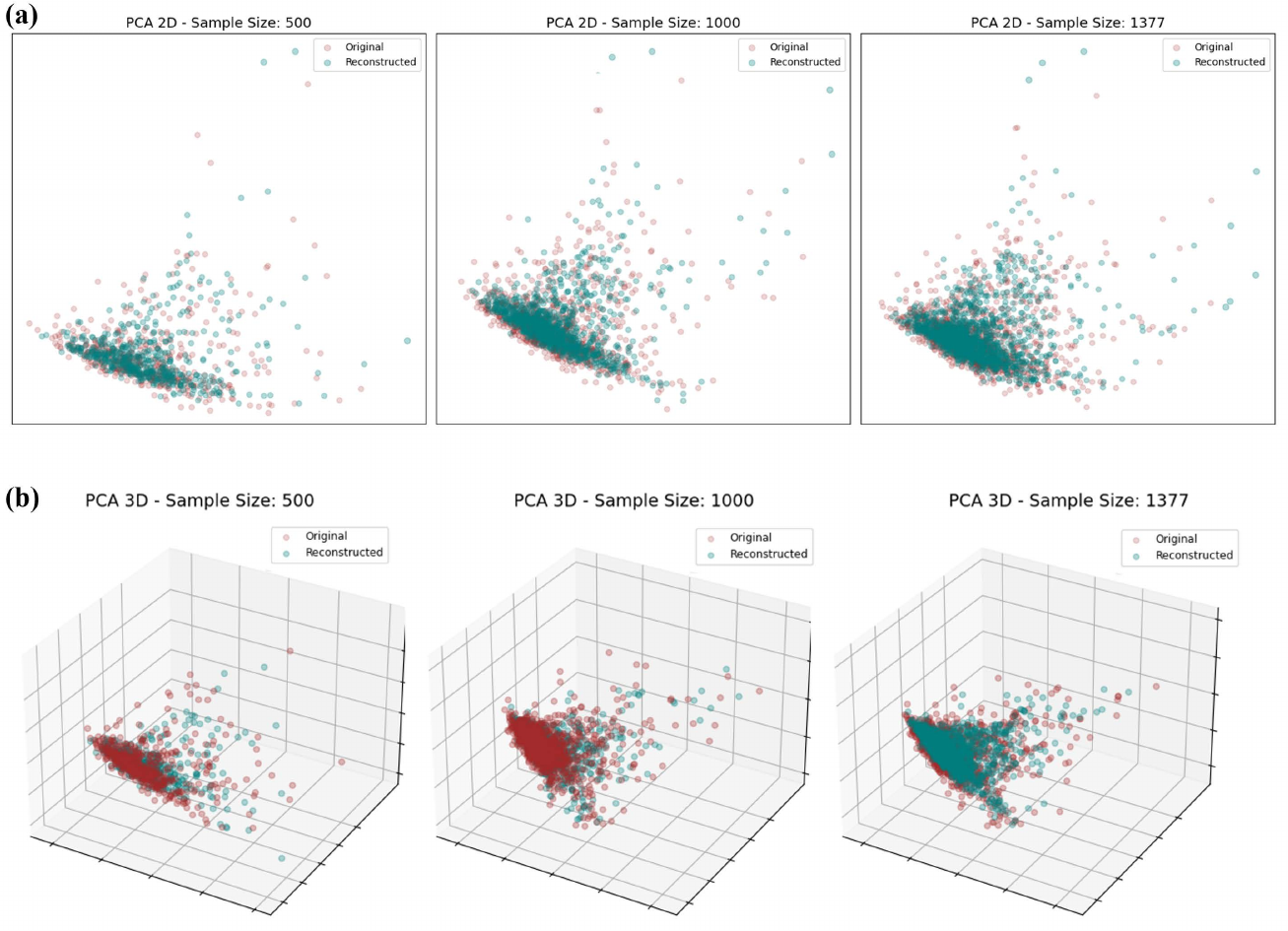}
\caption{PCA comparison between original and reconstructed omics expressions on the test set.}
\label{fig:vae-pca}
\vspace*{-0.5\baselineskip} 
\end{figure}
To assess whether the proposed \texttt{OmicsEn} module can effectively capture and reconstruct biologically meaningful features from gene expression profiles, we visualize the distributional similarity between the original and reconstructed omics representations. Specifically, we project both the input and reconstructed expression profiles into a shared PCA space for qualitative comparison. As shown in Figure \ref{fig:vae-pca}, we present both 2D (a) and 3D (b) PCA projections under varying sample sizes (500, 1000, and 1377). Across all settings, the reconstructed samples (green) closely overlap with the original ones (red), demonstrating that the latent encoding captures the essential structure of gene expression data. This confirms that \texttt{OmicsEn} preserves the biological fidelity of omics inputs, which is critical for guiding downstream molecular generation.

\subsubsection{Text-Guided Molecular Generation.}
\renewcommand{\thetable}{F.\arabic{table}} 
\renewcommand{\thefigure}{F.\arabic{figure}}
\renewcommand{\thealgorithm}{F.\arabic{algorithm}}
\label{appendix_F.2}
\paragraph*{Objective (1).}  
To quantitatively evaluate the alignment between generated molecules and their corresponding textual descriptions, we introduce the metric hit ratio. This metric assesses the extent to which a generated molecule contains the functional groups explicitly described in the input text prompt. Specifically, we extract the target functional group from each textual description and compute the noise estimation error between this reference and each generated sample. A molecule is considered a ``hit" if it exhibits a noise estimation error below a predefined threshold, which is set as the average noise error across all generated samples. The hit ratio is then calculated as the proportion of generated molecules that qualify as hits under this criterion. This evaluation serves as a functional-level validation of semantic consistency, complementing structure-based metrics such as BLEU or Tanimoto similarity. Higher hit ratios indicate stronger fidelity in capturing chemically meaningful semantics embedded in the text, validating the model’s ability to translate language-based prompts into structurally relevant molecular outputs.

%===============================================
\begin{table}[t]
\caption{Benchmark results of text-to-molecular generation on ChEBI-20 test split. We bold the best scores and underline the second-best scores.  \texttt{ToDi} \textsubscript{\scriptsize w/o O} denotes generation results guided solely by textual descriptions.}
\label{table:text}
\centering
\scalebox{0.9}{
\renewcommand{\arraystretch}{1.2}
\begin{tabular}{l|ccc|ccccc}\toprule
Model & Validity$\uparrow$ & BLEU$\uparrow$ & Levenshtein$\downarrow$ & MACCS$\uparrow$ & RDK$\uparrow$ & Morgan$\uparrow$ & FCD$\downarrow$ \\\midrule
Transformer   & \underline{0.91} & 0.50 & 1.15 & 0.48 & 0.32 & 0.22 & 0.23 \\
T5-Base   & 0.66 & 0.76 & 0.50 & 0.73 & 0.61 & 0.55 & 0.05 \\
MolT5-Base   & 0.77 & 0.77 & 0.49 & 0.72 & 0.59 & 0.53 & 0.04 \\
TGM   & 0.87 & \underline{0.83} & \textbf{0.34} & \underline{0.85} & \underline{0.74} & \textbf{0.69} & \textbf{0.01} \\\midrule
ToDi\textsubscript{\scriptsize w/o O}   & \textbf{1.00} & \textbf{0.89} & \underline{0.45} & \textbf{0.87} & \textbf{0.78} & \underline{0.63} & \underline{0.02} \\\bottomrule
\end{tabular}
}
\end{table}
%==========================================
\begin{figure}[t]
\centering
\includegraphics[width=0.9\linewidth]{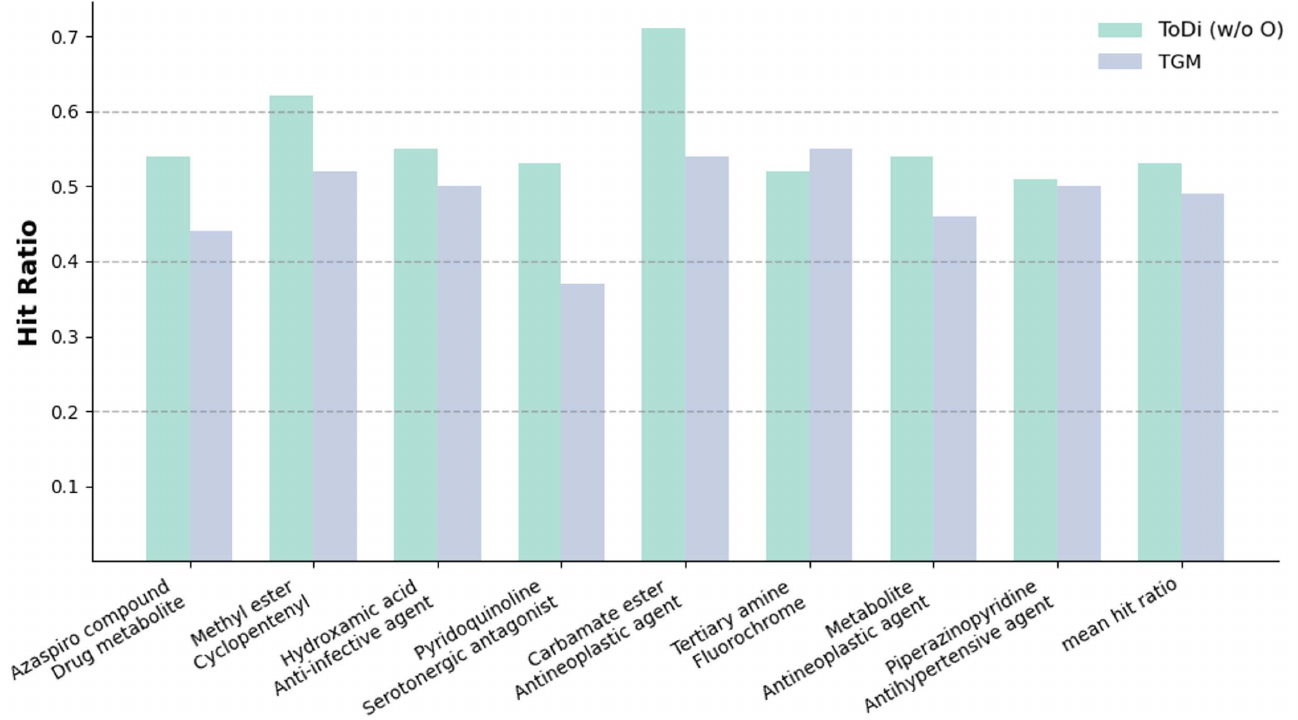}
\caption{Hit ratio comparison between \texttt{ToDi}\textsubscript{\scriptsize w/o O} and TGM on selected textual descriptions.}
\label{fig:hit-ratio-desc}
\vspace*{-0.5\baselineskip} 
\end{figure}
%==========================================
\paragraph*{Objective (2).}
Table~\ref{table:text} presents the benchmark results of text-to-molecular generation on the ChEBI-20 test split. \texttt{ToDi}\textsubscript{\scriptsize w/o O} achieves superior performance across most evaluation metrics, attaining the highest scores in Validity, BLEU, MACCS, and RDK. These results reflect the model’s strong capability in generating chemically valid structures that are both semantically faithful to the textual input and structurally consistent with reference molecules.
Notably, although \texttt{ToDi}\textsubscript{\scriptsize w/o O} ranks second in Levenshtein distance, Morgan similarity, and FCD, the margins are relatively small, and the model maintains competitive results across all metrics.
Taken together, these outcomes demonstrate the robustness of \texttt{ToDi}\textsubscript{\scriptsize w/o O} in capturing both fine-grained chemical semantics and global structural patterns, solely under the guidance of molecular textual descriptions. This validates its effectiveness in text-conditioned generation tasks, especially in settings where omics information is unavailable.
%====================================

\paragraph*{Fine-Grained Hit Ratio.} 
To further validate the semantic alignment between textual descriptions and generated molecules, we conduct a fine-grained analysis based on a subset of randomly selected textual descriptions. For each sentence, we extract two representative functional group terms and compute the hit ratio under the text-only setting. As shown in Figure \ref{fig:hit-ratio-desc}, \texttt{ToDi}\textsubscript{\scriptsize w/o O} consistently outperforms TGM across most textual scenarios, with noticeable gains in categories such as ``carbamate ester", ``methyl ester", and ``hydroxamic acid". The experiment is repeated 10 times to ensure statistical stability. The final bar on the right shows the mean hit ratio across all evaluated descriptions, where \texttt{ToDi}\textsubscript{\scriptsize w/o O} achieves a higher overall score than TGM. These results highlight the superior capacity of our method to align molecular generation with textual semantics, even without omics guidance.

%=============================================
\paragraph*{Correlation Evaluation.}  

\begin{figure}[t]
\centering
\includegraphics[width=0.9\linewidth]{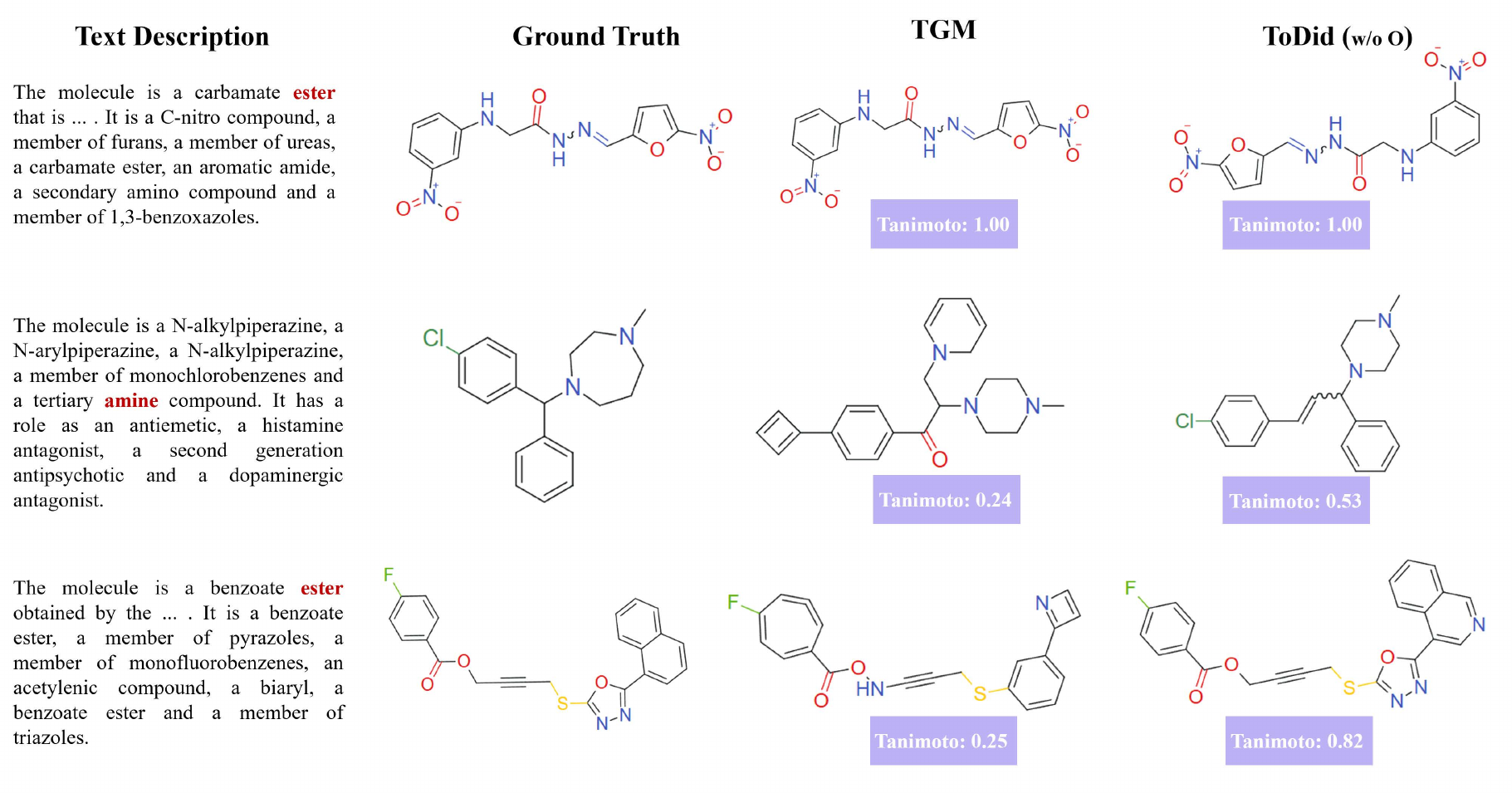}
\caption{Comparison of correlation results between \texttt{ToDi}\textsubscript{\scriptsize w/o O} and TGM on textual descriptions.}
\label{fig:text_todi}
\vspace*{-0.5\baselineskip} 
\end{figure}

To further demonstrate the semantic learning capability of \texttt{ToDi}, we visualize three representative cases in Figure \ref{fig:text_todi}. Compared to the state-of-the-art baseline TGM, \texttt{ToDi} consistently generates molecular structures that are more semantically aligned with the input descriptions. Notably, \texttt{ToDi} successfully captures and reconstructs key chemical functional groups explicitly mentioned in the textual input—such as \textit{ester}, \textit{amine}, and \textit{benzoate ester}—that TGM either misses or fails to represent accurately.

From a quantitative perspective, \texttt{ToDi} achieves Tanimoto similarity scores of 1.00, 0.53, and 0.82 across the three cases, outperforming TGM’s corresponding scores of 1.00, 0.24, and 0.25. The substantial improvements in the latter two examples (with gains of +0.29 and +0.57, respectively) highlight \texttt{ToDi}’s superior ability to perform fine-grained semantic-to-structure mapping and generate chemically faithful hit-like molecules.
%=========================================

\subsubsection{TextOmics-Guided Molecular Generation}
\label{appendix_F.3}
%==================================
\paragraph*{Extended Evaluation Metrics.} 
\begin{table}[t]
\centering
\setlength{\tabcolsep}{3pt}
\caption{Additional comparison of other metrics on ten target proteins between the \texttt{ToDi} method and the current SOTA methods.}
\resizebox{0.7\textwidth}{!}{
\begin{tabular}{l|ccc|ccc|ccc}\toprule
\multirow{2}{*}{Target} & \multicolumn{3}{c|}{BLEU$\uparrow$} & \multicolumn{3}{c|}{RDK$\uparrow$} & \multicolumn{3}{c}{QED$\uparrow$}\\
 & GxVAEs & TGM & \texttt{ToDi} & GxVAEs & TGM & \texttt{ToDi} & GxVAEs & TGM & \texttt{ToDi}\\
\midrule
AKT1   & 0.51 & 0.45 & \textbf{0.52} & 1.00 & 0.52 & \textbf{1.00} & 0.59 & 0.59 & \textbf{0.62}\\
AKT2   & 0.52 & 0.47 & \textbf{0.53} & 0.66 & 0.51 & \textbf{0.83} & 0.57 & 0.60 & \textbf{0.62}\\
AURKB  & 0.50 & 0.47 & \textbf{0.52} & 0.70 & 0.58 & \textbf{0.81} & 0.58 & 0.61 & \textbf{0.62}\\
CTSK   & 0.45 & 0.46 & \textbf{0.50} & 0.60 & 0.49 & \textbf{0.89} & 0.61 & 0.58 & \textbf{0.62}\\
EGFR   & 0.42 & 0.47 & \textbf{0.52} & 0.80 & 0.75 & \textbf{0.98} & 0.62 & 0.61 & \textbf{0.63}\\
HDAC1  & 0.50 & 0.48 & \textbf{0.52} & 0.66 & 0.64 & \textbf{0.83} & 0.59 & 0.60 & \textbf{0.61}\\
MTOR   & 0.48 & 0.48 & \textbf{0.50} & 0.61 & 0.57 & \textbf{0.89} & \textbf{0.68} & 0.60 & \underline{0.61}\\
PIK3CA & 0.46 & 0.45 & \textbf{0.49} & 0.55 & 0.52 & \textbf{0.72} & 0.66 & 0.60 & \textbf{0.66}\\
SMAD3  & 0.56 & 0.48 & \textbf{0.58} & 0.92 & 0.81 & \textbf{1.00} & 0.61 & 0.60 & \textbf{0.62}\\
TP53   & 0.46 & 0.44 & \textbf{0.59} & 0.86 & 0.83 & \textbf{1.00} & 0.60 & 0.61 & \textbf{0.62}\\\bottomrule
\end{tabular}
}
\label{tab:extra-metrics}
\end{table}
%=============================

To provide a more comprehensive evaluation of \texttt{TextOmics}-conditioned molecular generation, we compare \texttt{ToDi} with two representative baseline methods, GxVAEs and TGM, across three additional evaluation metrics on 10 target proteins, as summarized in Table~\ref{tab:extra-metrics}. In addition to BLEU and RDK similarity—which respectively assess the sequence-level and fingerprint-level alignment between generated and ground-truth compounds—we incorporate the Quantitative Estimate of Drug-likeness (QED), a widely adopted metric that quantifies the likelihood of a molecule possessing favorable pharmacokinetic and physicochemical properties.

QED is computed as a weighted combination of several molecular descriptors, including molecular weight, octanol–water partition coefficient (logP), the number of hydrogen bond donors and acceptors, and the number of rotatable bonds. This provides a holistic measure of a compound’s suitability as a potential drug candidate. Across the majority of targets, \texttt{ToDi} consistently achieves the highest scores in all three metrics, with particularly strong performance in RDK and QED, reflecting both structural fidelity and drug-like characteristics. Although \texttt{ToDi} ranks second in QED for the MTOR target, the difference is marginal, and it maintains the best overall drug-likeness profile across the benchmark.

\begin{figure}[t]
\centering
\includegraphics[width=1.0\textwidth]{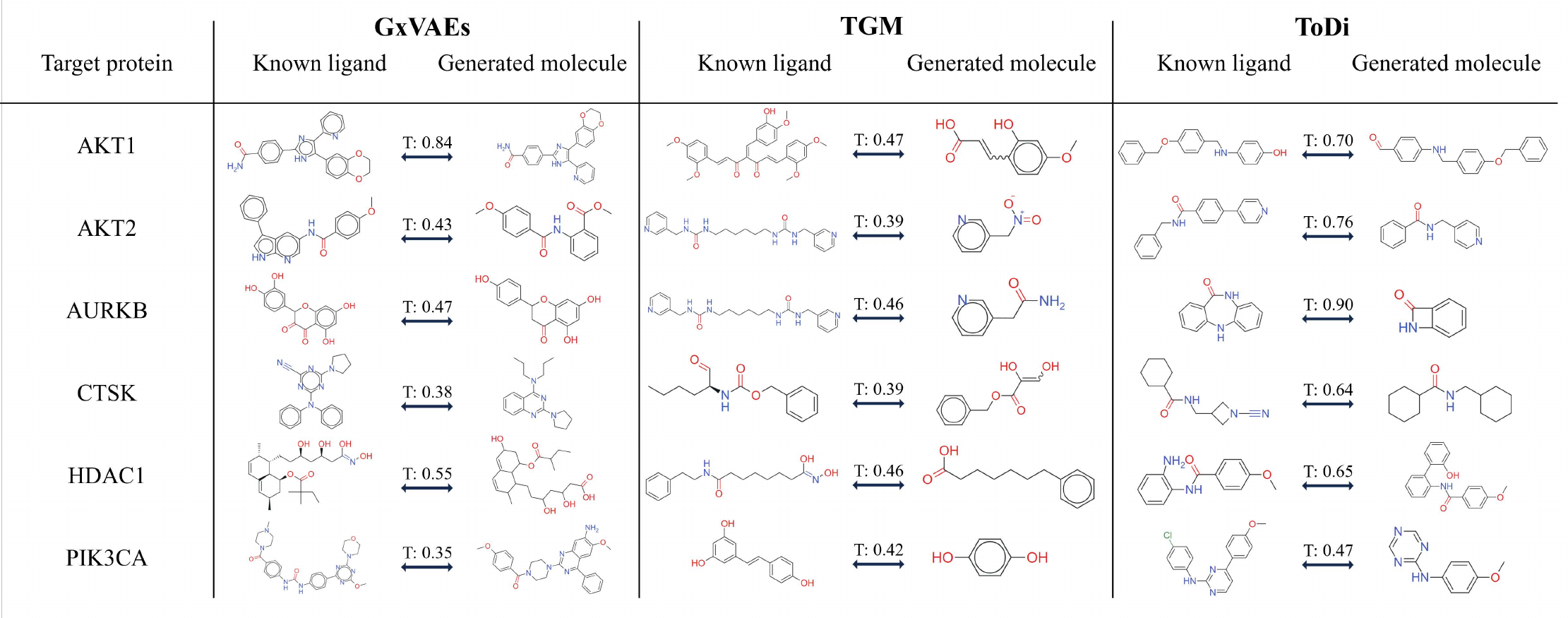}
\caption{Detailed illustration of \texttt{ToDi}'s performance in terms of biological similarity and textual hit under both overexpression and knockdown conditions.}
\label{fig:ligands}
\vspace*{-1\baselineskip} 
\end{figure}

These results collectively highlight \texttt{ToDi}’s capacity to generate molecules that are not only semantically consistent with textual prompts but also chemically and pharmacologically meaningful. The remaining comparison results of Morgan Tanimoto similarity for overexpressed molecules against baseline models are provided in Figure~\ref{fig:ligands}, further supporting the structural relevance of the generated compounds.
%==========================================

%=============================================================
\end{document}